\documentclass{article}
\pdfoutput=1
\usepackage[main,final]{neurips_2026}
\usepackage{natbib}
\setcitestyle{numbers}
\setcitestyle{square}
\setcitestyle{comma}
\usepackage{multirow}
\usepackage[utf8]{inputenc} 
\usepackage[T1]{fontenc}    
\usepackage{url}            
\usepackage{booktabs}       
\usepackage{amsfonts}       
\usepackage{nicefrac}       
\usepackage{microtype}      
\usepackage{xcolor}         
\usepackage{algorithm}
\usepackage{algorithmic}
\usepackage{bm}
\usepackage{lineno} 
\usepackage{makecell}
\usepackage{amsthm}
\usepackage{amssymb}
\usepackage{amsmath}
\usepackage{wrapfig}
\usepackage{graphicx}
\usepackage{float}
\usepackage{subfigure}
\theoremstyle{plain}
\usepackage{varwidth}
\usepackage{dsfont}

\usepackage{ulem}
\usepackage{colortbl,color}
\usepackage[hidelinks]{hyperref}
\definecolor{citeColor}{RGB}{0,20,115}
\hypersetup{colorlinks,linkcolor={citeColor},citecolor={citeColor},urlcolor={citeColor}}
\definecolor{LightGray}{HTML}{F0F1F2} 

\usepackage{pifont}
\usepackage{tikz}
\usetikzlibrary{arrows.meta,positioning,calc,fit,backgrounds}
\usetikzlibrary{arrows.meta,calc,positioning}
\usepackage{pgfplots}
\pgfplotsset{compat=1.18}
\usetikzlibrary{patterns.meta}
\definecolor{chartSH}{HTML}{287FA3}   
\definecolor{chartGrayA}{HTML}{E4E6E9}
\definecolor{chartGrayB}{HTML}{C6CBD1}
\definecolor{chartGrayC}{HTML}{A7ADB6}
\definecolor{chartGrayD}{HTML}{8A919B}
\definecolor{chartBaseRef}{HTML}{B8BEC6}
\newcommand{\chartkey}[2]{\textcolor{#1}{\rule{7pt}{4.2pt}}\,\,#2}
\pgfplotsset{
  chartbase/.style={
    ybar, scale only axis, axis lines*=left, ymajorgrids, grid style={black!8},
    major tick length=0pt, axis line style={black!40},
    tick label style={font=\sffamily\fontsize{6}{7}\selectfont, black},
    label style={font=\sffamily\fontsize{6}{7}\selectfont, black},
    title style={font=\sffamily\bfseries\fontsize{7}{8}\selectfont, black},
    clip=false, point meta=y, /pgf/number format/assume math mode=true,
  },
  chartvalue/.style={
    nodes near coords, every node near coord/.append style={
      font=\sffamily\fontsize{4.8}{5.5}\selectfont, xshift=-2.6pt, yshift=1.4pt, rotate=60, anchor=west, inner sep=0pt, text=black,
      /pgf/number format/fixed, /pgf/number format/fixed zerofill, /pgf/number format/precision=2},
  },
  chartvaluesh/.style={every node near coord/.append style={font=\sffamily\bfseries\fontsize{4.8}{5.5}\selectfont}},
  chartvalueflat/.style={
    nodes near coords, every node near coord/.append style={
      font=\sffamily\fontsize{4.5}{5.5}\selectfont, anchor=south, inner sep=.8pt, text=black!60,
      /pgf/number format/fixed, /pgf/number format/fixed zerofill, /pgf/number format/precision=2},
  },
  chartvalueflatsh/.style={every node near coord/.append style={font=\sffamily\bfseries\fontsize{4.5}{5.5}\selectfont, text=chartSH}},
  chartsh/.style={fill=chartSH, draw=chartSH},
  chartsolid/.style={fill=#1, draw=#1},
  charthatch/.style={fill=#1, draw=#1, postaction={pattern={Lines[angle=45,distance=3.2pt,line width=1pt]}, pattern color=white}},
  chartlegend/.style={
    legend style={at={(0.5,1.03)}, anchor=south, draw=none, fill=none, inner sep=1pt, legend columns=#1,
      font=\sffamily\fontsize{6}{7}\selectfont, text=black, /tikz/every even column/.style={column sep=5pt}},
    legend cell align=left, legend image code/.code={\draw[##1] (0pt,-1.7pt) rectangle (9pt,3.3pt);},
  },
}

\definecolor{deepgreen}{RGB}{0,128,0}
\definecolor{myred}{RGB}{200,50,50}
\definecolor{symgray}{RGB}{70,70,70}

\definecolor{py-keyword}{RGB}{0,0,255}       
\definecolor{py-string}{RGB}{163,21,21}      
\definecolor{py-comment}{RGB}{0,128,0}       
\definecolor{py-builtin}{RGB}{0,112,193}     
\definecolor{py-bg}{HTML}{FFFFFF}            
\definecolor{py-fg}{HTML}{000000}            
\definecolor{api-bg}{HTML}{F6F8FA}            
\definecolor{api-fn}{HTML}{0B4F9C}            
\definecolor{api-comment}{HTML}{6A737D}       
\definecolor{py-lineno}{HTML}{858585}        

\usepackage{xspace}

\usepackage{array}
\usepackage{enumerate}
\usepackage{tabularx}
\usepackage{enumitem}
\usepackage{bbding}
\usepackage{caption}
\usepackage[toc,page,header]{appendix}
\usepackage{sidecap}
\usepackage[most]{tcolorbox}
\renewcommand{\emph}[1]{\textit{#1}}

\theoremstyle{plain}

\newcolumntype{L}[1]{>{\raggedright\let\newline\\\arraybackslash\hspace{0pt}}m{#1}}
\newcolumntype{C}[1]{>{\centering\let\newline  \\\arraybackslash\hspace{0pt}}m{#1}}%
\newcolumntype{R}[1]{>{\raggedleft\let\newline \\\arraybackslash\hspace{0pt}}m{#1}}

\definecolor{shadecolor}{rgb}{0.92,0.92,0.92}

\newcolumntype{Y}{>{\raggedright\arraybackslash}X}

\usepackage[capitalize,noabbrev]{cleveref}

\usepackage{etoc}
\etocdepthtag.toc{mtchapter}
\etocsettagdepth{mtchapter}{subsection}
\etocsettagdepth{mtappendix}{none}

\newcounter{finding}
\newcommand{\findingbox}[2][]{%
  \refstepcounter{finding}%
  \begin{tcolorbox}[
    colback=blue!6!white,
    colframe=blue!45!black,
    boxrule=0.5pt,
    arc=3pt,
    left=8pt, right=8pt, top=6pt, bottom=6pt,
  ]
  \textbf{Finding \thefinding.}%
  \if\relax\detokenize{#1}\relax
  \else
    \label{#1}%
  \fi
  \enspace #2
  \end{tcolorbox}
}

\title{
Spatial Harness: Grounding Spatial Reasoning\\
in Executable World States with Physics Engines
}

\author{%
  Chentao Cao\textsuperscript{1}\thanks{Equal contribution.}\quad
  Zhanke Zhou\textsuperscript{1}\footnotemark[1]\quad
  Sangni Duan\textsuperscript{1}\quad
  Bo Han\textsuperscript{1}\thanks{Correspondence to Bo Han (bhanml@comp.hkbu.edu.hk).}\quad
  Hang Li\textsuperscript{2}\\
  \textsuperscript{1}TMLR Group, Department of Computer Science, Hong Kong Baptist University\\
  \textsuperscript{2}ByteDance Seed
}

\begin{document}

\maketitle

\begin{abstract}
Spatial reasoning requires understanding how objects and observers are arranged in a scene and how their relations change with viewpoint or movement.
Vision-language models (VLMs) typically perform both steps through \emph{model-internal} reasoning, forming an \emph{implicit} scene interpretation from partial views and imagining the consequences of spatial transformations.
A misread scene can lead to a wrong answer, while even a correct interpretation does not ensure that imagined transformations satisfy spatial constraints.
We introduce \emph{Spatial Harness}, which represents scenes as explicit and executable world states to ground spatial reasoning in physics-engine feedback.
A Perceiver integrates scene constraints extracted from the question with specialized visual estimates to place objects and cameras in a shared coordinate frame.
The resulting state is instantiated in a physics engine, where a Reasoner executes viewpoint changes and uses computed measurements and rendered views to guide reasoning.
We evaluate on MindCube, SAT, MMSI-Bench, and SpatialRGPT-Bench with open- and closed-weight VLMs.
On MindCube, Spatial Harness improves Qwen3.5-27B from 52.70\% to 71.14\%, surpassing pySpatial by 7.52 percentage points.
Given identical annotated object and camera poses, it reaches 98.25\% versus 71.81\% for model-internal reasoning.
Code is publicly available at: \url{https://github.com/tmlr-group/Spatial-Harness}.
\end{abstract}

\begin{figure}[H]
\centering
\includegraphics[width=\textwidth]{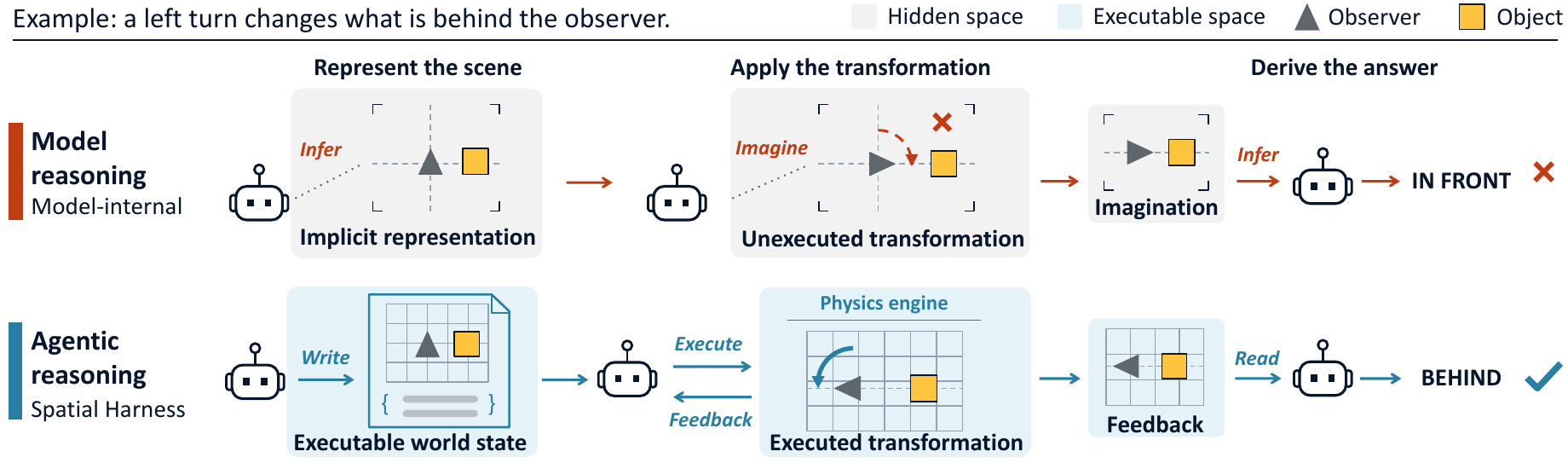}
\caption{\textbf{From model-internal reasoning to executable world states.} Both approaches proceed from scene representation through spatial transformation to answer derivation. Top: a VLM implicitly represents the scene and imagines the requested turn, leaving the resulting spatial relation unchecked. Bottom: Spatial Harness constructs an explicit world state and executes the turn in a physics engine, grounding the answer in computed feedback. This makes both the estimated layout and the consequences of the transformation available for inspection. Execution enforces consistency with this state, whose accuracy still depends on the available language constraints and visual estimates.}
\label{fig:motivation}
\end{figure}

\begin{figure}[H]
\centering
\includegraphics[width=\textwidth]{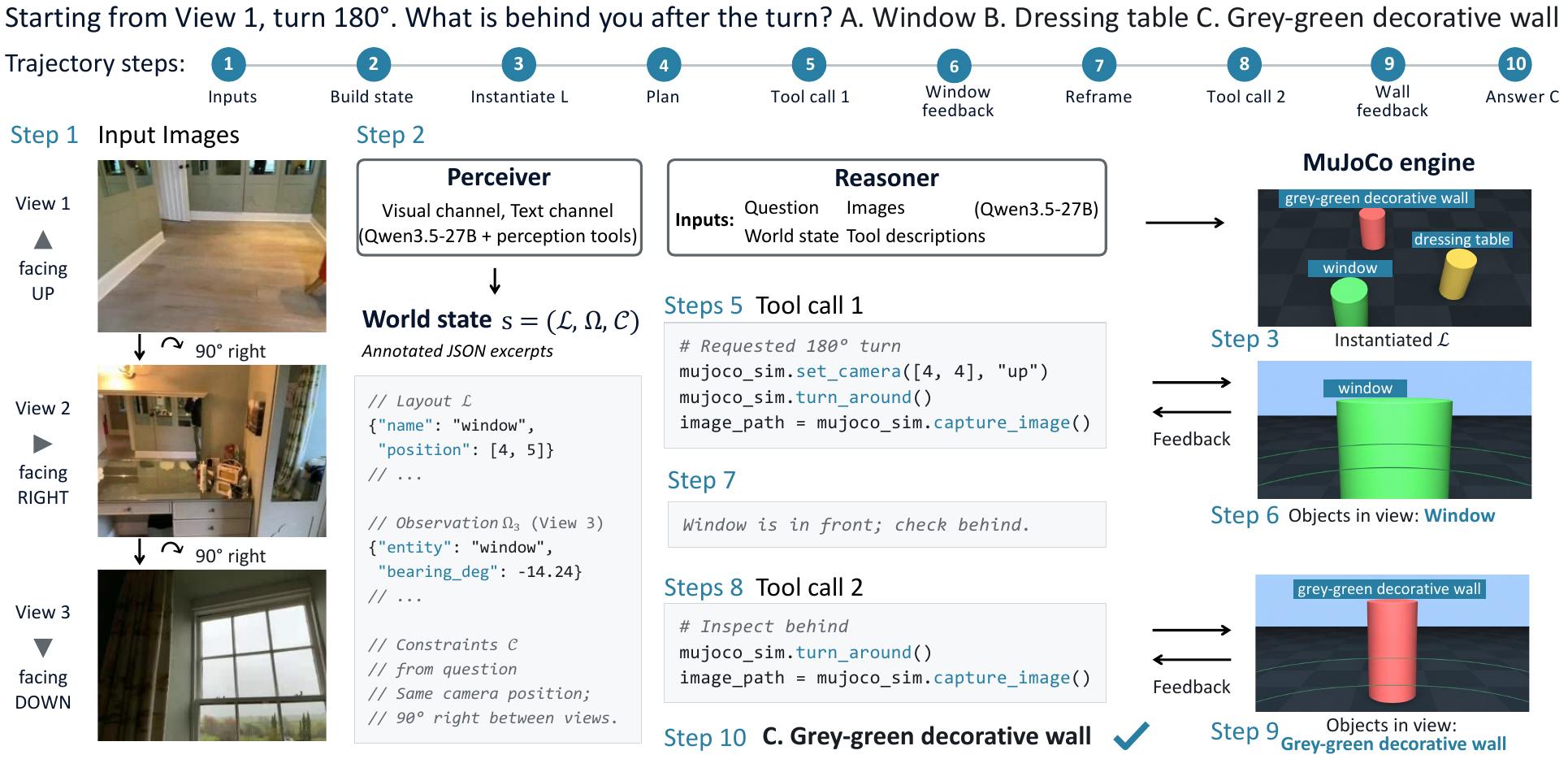}
\caption{\textbf{Spatial reasoning through iterative execution and feedback.} Given three views linked by $90^\circ$ camera rotations, the Perceiver constructs a shared world state and instantiates it in MuJoCo (steps 1--3). The Reasoner executes the requested $180^\circ$ turn and renders the new view (steps 4--6). The window is now in front, but the question asks what is behind. The Reasoner therefore turns again to inspect the opposite direction, identifies the grey-green decorative wall, and selects C (steps 7--10).}
\label{fig:demo}
\vspace{-.1in}
\end{figure}


\section{Introduction}
\label{sec:introduction}
Spatial reasoning requires vision-language models (VLMs) to infer spatial relations beyond those directly observed in the input views~\citep{yin2025spatial,ray2024sat}.
Determining what lies behind an observer after a turn, for example, entails recovering the scene layout and evaluating object relations in the transformed reference frame.
Despite advances in multimodal reasoning~\citep{suris2023vipergpt,zheng2026deepeyes,hong2026deepeyesv,zhang2026thyme}, this remains challenging: GPT-5 achieves 47.59\% accuracy on MindCube, compared with 32.35\% for random guessing~\citep{yin2025spatial}.

The difficulty is that direct reasoning leaves both \emph{scene representation} and \emph{spatial transformation} within the model (Figure~\ref{fig:motivation}, top).
A VLM may infer an incorrect layout and reason consistently from it, or infer the layout correctly but misapply the requested transformation.
The final answer alone cannot distinguish these failures, and a fluent explanation provides no independent check of either step.
Additional views, as generated by MindJourney~\citep{yang2025mindjourney}, provide visual evidence but do not by themselves enforce consistency between scene estimates and subsequent spatial transformations.

Our central premise is that a scene representation should determine the consequences of spatial operations.
An explicit \emph{world state} makes estimated objects, poses, and relations inspectable, but providing these facts as text still leaves their consequences to model-internal reasoning.
Making the state \emph{executable} instead allows an engine to compute measurements and render views from the specified geometry (Figure~\ref{fig:motivation}, bottom).
Successive operations then share a common scene, against which the model can check its spatial judgments.
Execution does not eliminate perceptual uncertainty, but it makes the assumed layout and the results of operations separately available for inspection.

We introduce \emph{Spatial Harness}, which implements this principle with a Perceiver and a Reasoner in \mbox{MuJoCo}~\citep{todorov2012mujoco}.
The Perceiver integrates specialized visual estimates with constraints stated in the question, including camera movements that relate observations across views, to place objects and cameras in a shared coordinate frame.
The Reasoner queries the instantiated state through Python, using viewpoint changes, geometric measurements, and rendered views to resolve the question.
Language constraints and visual geometry jointly define the scene used for spatial computation.

Figure~\ref{fig:demo} presents a concrete example: determining what lies behind an observer after a $180^\circ$ turn.
After executing the turn, the first rendering shows a window ahead, whereas the question asks what is behind.
The Reasoner therefore performs another $180^\circ$ turn to inspect that direction, identifies the grey-green decorative wall, and selects C.
The sequence illustrates how \emph{execution feedback} guides subsequent queries while the engine preserves a shared world state throughout the reasoning process.

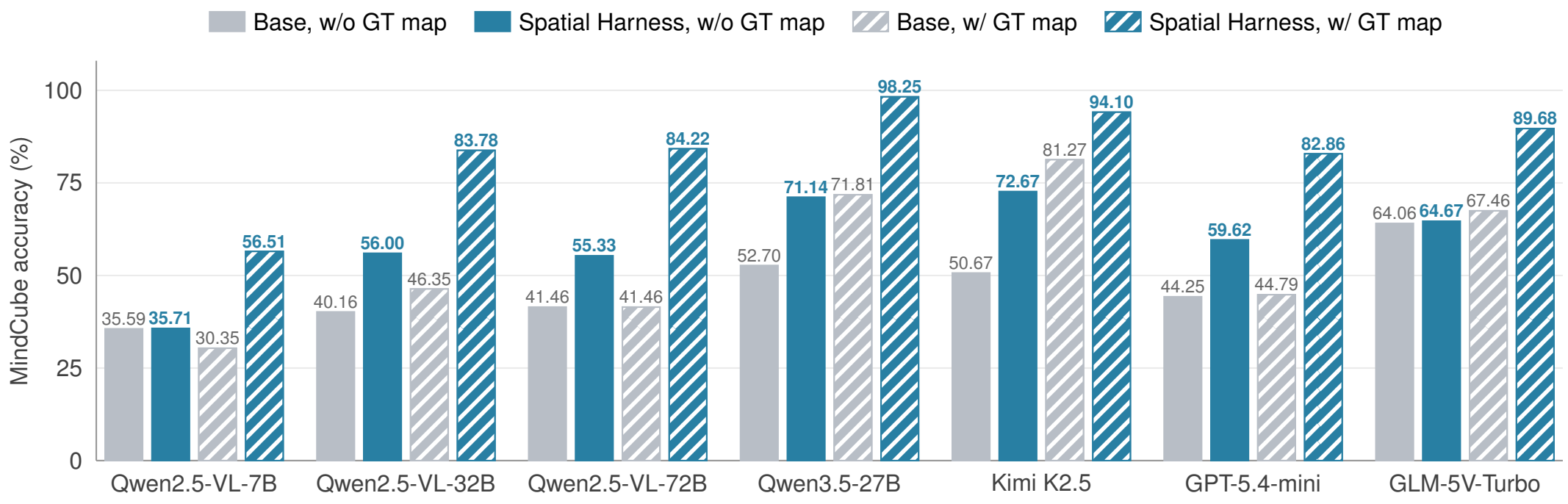
\begin{figure}[!t]
\centering
\begin{tikzpicture}
\begin{axis}[
  chartbase, ybar=2.4pt, chartvalueflat, chartlegend=4, legend style={at={(0.5,1.03)}, anchor=south}, bar width=9.6pt, width=0.945\textwidth, height=3.6cm, ymin=0, ymax=108, ytick={0,25,50,75,100},
  tick label style={font=\sffamily\fontsize{6}{7}\selectfont, black!75}, label style={font=\sffamily\fontsize{6}{7}\selectfont, black!75},
  ylabel={MindCube accuracy (\%)}, ylabel style={yshift=-4pt},
  symbolic x coords={m7B,m32B,m72B,m27B,mKimi,mGPT,mGLM}, xtick={m7B,m32B,m72B,m27B,mKimi,mGPT,mGLM}, enlarge x limits=0.077,
  xticklabels={Qwen2.5-VL-7B, Qwen2.5-VL-32B, Qwen2.5-VL-72B, Qwen3.5-27B, Kimi K2.5, GPT-5.4-mini, GLM-5V-Turbo},
]
\addplot+[chartsolid=chartBaseRef] coordinates {(m7B,35.59) (m32B,40.16) (m72B,41.46) (m27B,52.70) (mKimi,50.67) (mGPT,44.25) (mGLM,64.06)};
\addplot+[chartsh, chartvalueflatsh] coordinates {(m7B,35.71) (m32B,56.00) (m72B,55.33) (m27B,71.14) (mKimi,72.67) (mGPT,59.62) (mGLM,64.67)};
\addplot+[charthatch=chartBaseRef] coordinates {(m7B,30.35) (m32B,46.35) (m72B,41.46) (m27B,71.81) (mKimi,81.27) (mGPT,44.79) (mGLM,67.46)};
\addplot+[charthatch=chartSH, chartvalueflatsh] coordinates {(m7B,56.51) (m32B,83.78) (m72B,84.22) (m27B,98.25) (mKimi,94.10) (mGPT,82.86) (mGLM,89.68)};
\legend{{Base, w/o GT map}, {Spatial Harness, w/o GT map}, {Base, w/ GT map}, {Spatial Harness, w/ GT map}}
\end{axis}
\end{tikzpicture}
\vspace{-.12in}
\caption{\textbf{Separating scene information from its use in reasoning.} MindCube accuracy across seven VLMs. Gray and blue bars denote direct reasoning (Base) and Spatial Harness, respectively. Solid bars use images and questions without a ground-truth (GT) map, while hatched bars supply both methods with the same GT layout. With identical GT layouts, Spatial Harness outperforms \textit{Base} on every backbone, demonstrating gains from executing the supplied layout.}
\label{fig:mindcube-gt-comparison}
\vspace{-.1in}
\end{figure}

Holding the world state fixed allows us to evaluate reasoning independently of scene estimation.
Figure~\ref{fig:mindcube-gt-comparison} compares \textit{Base} (the same VLM answering without external tools) and Spatial Harness with and without ground-truth (GT) maps specifying object and camera positions and headings.
GT maps do not consistently improve \textit{Base}, but Spatial Harness benefits from executing operations on the supplied state.
With Qwen3.5-27B, the same layout yields 71.81\% accuracy with \textit{Base} and 98.25\% with Spatial Harness, so the gain cannot be attributed to a better scene estimate.
Evaluations without GT maps cover MindCube~\citep{yin2025spatial}, SAT~\citep{ray2024sat}, MMSI-Bench~\citep{yang2026mmsi}, and SpatialRGPT-Bench~\citep{cheng2024spatialrgpt}. We also ablate state construction and execution feedback and assess estimated-state quality.

Our contributions are as follows:
\begin{itemize}[leftmargin=*,itemsep=2pt,topsep=3pt]
    \item
    \textbf{Executable world states.} We connect inspectable scene estimates to computed spatial consequences through an explicit interface for spatial reasoning grounded in geometry (Section~\ref{sec:method:overview}).
    \item
    \textbf{Spatial Harness.} We integrate language constraints and visual estimates into a shared world state and support interactive reasoning with measurements and rendered views (Sections~\ref{sec:method:perceiver}--\ref{sec:method:reasoner}).
    \item
    \textbf{Controlled evaluation.} We evaluate four benchmarks, compare reasoning with matched states (Figure~\ref{fig:mindcube-gt-comparison}), and analyze scene construction, execution feedback, and state quality (Sections~\ref{sec:experiments}--\ref{sec:discuss}).
\end{itemize}

\section{Related Work}
\label{sec:related-work}

\textbf{Agentic Multimodal Reasoning.}
ReAct, ViperGPT, and VisProg organize reasoning around external actions or executable programs~\citep{yao2022react,suris2023vipergpt,gupta2023visual}.
DeepEyes and THYME learn visual tool use~\citep{zheng2026deepeyes,hong2026deepeyesv,zhang2026thyme}, while agent frameworks develop planning and verification~\citep{lu2025octotools,zhou2025alphaapollo,zhou2026alphadiana}.
These approaches chiefly concern how tools are selected, invoked, and coordinated during reasoning.
Spatial Harness complements such orchestration by specifying the geometric state shared by spatial tools, so successive calls return measurements and views derived from the same explicit and inspectable scene assumptions.

\textbf{Spatial Reasoning in VLMs.}
SpatialVLM, Visual Spatial Tuning, and SpatialRGPT enhance spatial understanding through training or geometric inputs~\citep{chen2024spatialvlm,yang2025visual,cheng2024spatialrgpt}.
SpatialPIN and APC construct spatial priors or scene abstractions with visual experts~\citep{ma2024spatialpin,lee2025perspective}.
Among tool-based methods, pySpatial executes programs over reconstructed 3D geometry~\citep{luo2026pySpatial}, while GCA constrains computation by formalizing the query's reference frame and objective~\citep{chen2025geometrically}.
Spatial Harness uses question-stated object relations and camera motions to constrain the estimated world state itself.
These constraints determine the scene layout, rather than only the reference frame used to query it.

\textbf{Simulation-Grounded Reasoning.}
MindJourney synthesizes additional viewpoints with a video world model~\citep{yang2025mindjourney}.
Mind's Eye simulates physical outcomes from textual descriptions~\citep{liu2022mindseye}, and LLMPhy estimates physical parameters and layouts through simulation feedback~\citep{cherian2024llmphy}.
Spatial Harness uses a physics engine for \emph{spatial measurements and viewpoint rendering} over an estimated static scene.
Its world state specifies object and camera geometry, so these queries require neither physical parameter estimation nor prediction of object dynamics.

\clearpage
\section{Method}
\label{sec:methodology}

\begin{figure*}
\begin{center}
\includegraphics[width=\textwidth]{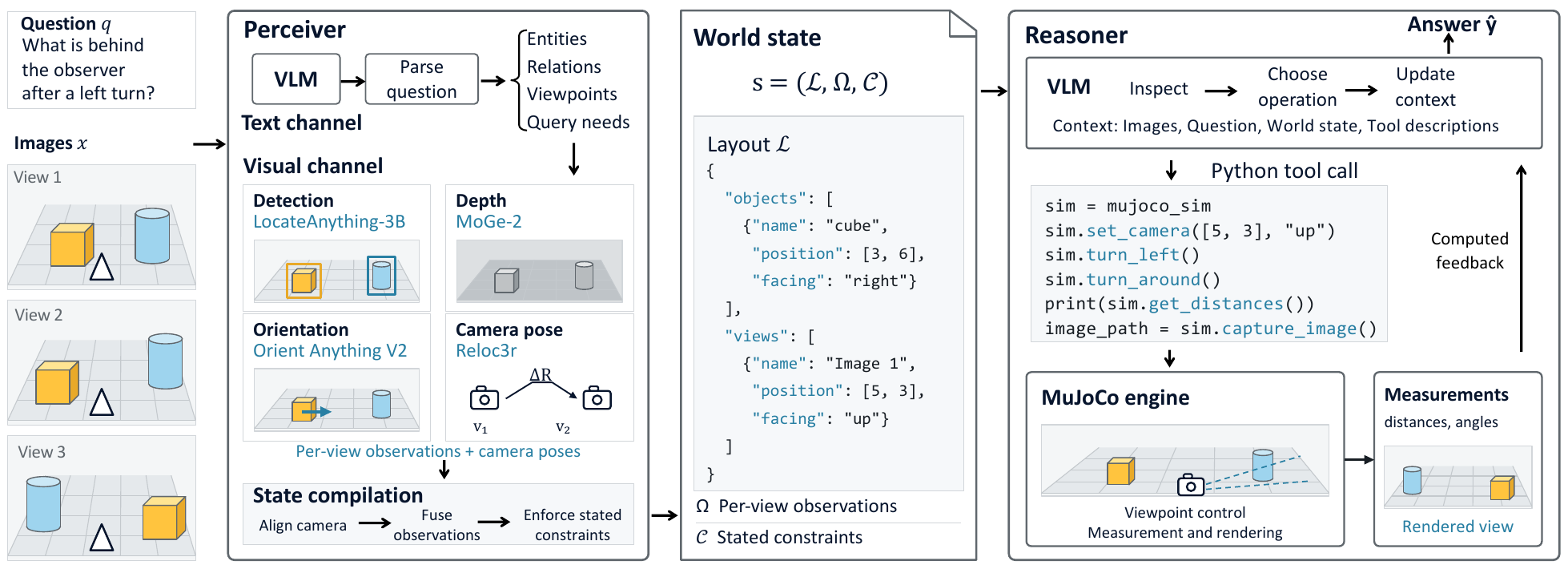}
\end{center}
\vspace{-.05in}
\caption{\textbf{Overview of Spatial Harness.} The Perceiver combines question-derived constraints with specialized visual estimates, aligning camera frames and fusing observations into a JSON world state $s=(\mathcal{L},\Omega,\mathcal{C})$: a shared layout, per-view observations, and stated constraints. The Reasoner instantiates $\mathcal{L}$ in MuJoCo and issues Python tool calls for viewpoint control, measurement, and rendering. Computed measurements and rendered views update its context and guide subsequent operations and the final answer. The JSON and code are illustrative excerpts.}
\vspace{-.15in}
\label{fig:architecture}
\end{figure*}


\subsection{Executable World States for Grounded Reasoning}
\label{sec:method:overview}

\textbf{Problem formulation.} We consider spatial question answering. Each instance consists of a set of images $x=\{I_i\}_{i=1}^{m}$ of a scene with $m\geq 1$, a textual query $q$ that poses the question and may describe part of the scene, and a ground-truth answer $y\in\mathcal{Y}$. A model takes $(x,q)$ and predicts $\hat{y}\in\mathcal{Y}$. Answering $q$ involves two steps: \emph{perception}, which establishes the scene facts that the question depends on, such as where the entities are and from where each image was taken, and \emph{reasoning}, which derives the answer from these facts, for example by working out how the scene looks from another viewpoint. A VLM that answers directly from $(x,q)$ carries out both steps within one generation. The perceived facts are not represented separately from the reasoning, so an error in perception cannot be told apart from an error in reasoning, and neither step can be checked or improved on its own. Spatial Harness separates the two steps: it makes the perceived facts explicit as a \emph{world state} and grounds the reasoning in computation on that state, so that every spatial judgment, such as what a viewpoint change brings into view, is computed by a physics engine from the state rather than produced by the VLM. The two steps can then be checked and improved separately.

\textbf{World state.} A \emph{world state} is an explicit representation of the scene facts that $q$ depends on. Let $\mathcal{O}=\{o_j\}_{j=1}^{n}$ denote the entities that $q$ involves; the state covers these entities and the properties the question depends on, namely their spatial configuration and that of the cameras, rather than the whole scene. We require the state to be \emph{executable}: it specifies the geometry concretely enough that geometric quantities, such as the distance between two entities, the direction of an entity from a chosen viewpoint, or the view from a camera placed anywhere in the scene, can be computed from it.

\begin{figure*}[t]
\begin{center}
\includegraphics[width=\textwidth]{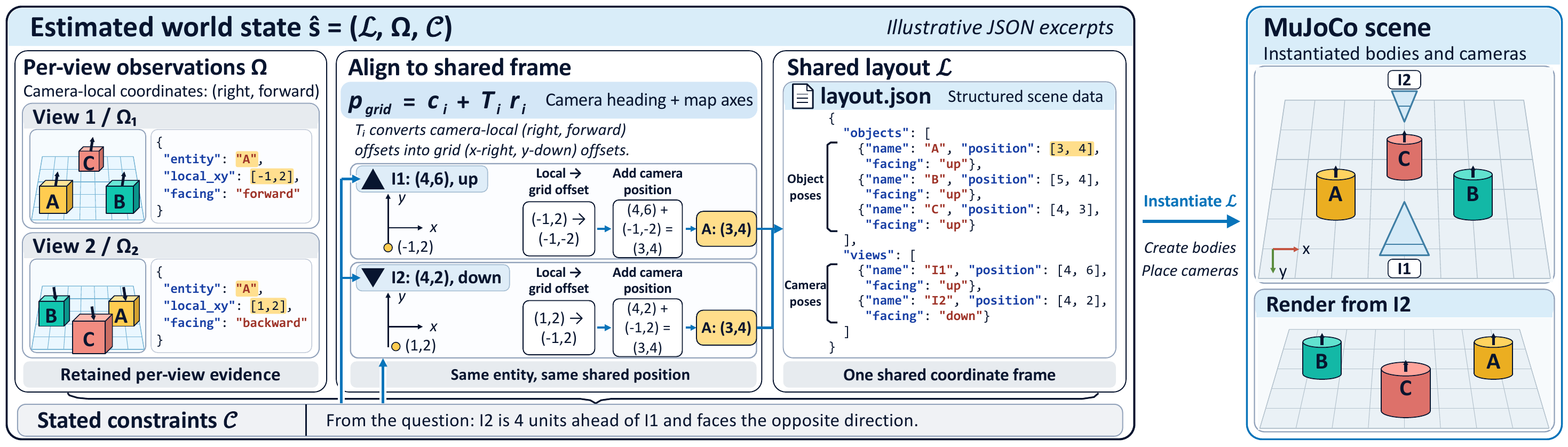}
\end{center}
\vspace{-.05in}
\caption{\textbf{World state and its engine instantiation.} Camera poses align per-view observations into a shared layout (Eq.~\ref{eq:shared-frame-transform}), instantiated as labeled proxies. For display, the first-camera origin is shifted to $(4,6)$, with $y$ increasing downward; $T_i$ combines camera rotation with this grid-axis convention.}
\vspace{-.2in}
\label{fig:world-state}
\end{figure*}

The world state has two levels: the observations from each image, and the arrangement of all entities and cameras in one scene (Figure~\ref{fig:world-state}). The \emph{observation level} holds what each image shows: $\Omega_i$ lists the entities visible in $I_i$ with their direction, distance, and facing relative to camera $i$, measured metrically or described qualitatively. Observations from different images are expressed relative to different cameras and can be related to each other only once the camera poses are known. The \emph{layout level} places all entities and cameras in a single coordinate system, which we call the shared frame: $\mathcal{L}$ gives the position and heading $(\mathbf{p}_j,\theta_j)$ of every entity and $(\mathbf{c}_i,\phi_i)$ of every camera, obtained by aligning the observations through the camera poses, and it is the part that is instantiated for computation. The \emph{constraints} $\mathcal{C}$ collect the scene facts stated in $q$, such as a relation between two entities or between two viewpoints; they hold exactly and apply to both levels. The observation level is kept alongside the layout because the alignment can lose or distort what an individual image shows, and the Reasoner can consult the observations directly. The world state is thus $s=(\mathcal{L},\Omega,\mathcal{C})$ with $\Omega=\{\Omega_i\}_{i=1}^{m}$.

\textbf{Decoupling perception from reasoning.} Spatial Harness consists of a Perceiver $\mathcal{P}$, which estimates the state from the images and the query, and a Reasoner $\mathcal{R}$, which answers the query by operating on the scene $\mathcal{E}[\hat{s}]$ that the engine instantiates from the estimate (Figure~\ref{fig:architecture}):
\begin{equation}
    \hat{s}=\mathcal{P}(x,q),
    \qquad
    \hat{y}=\mathcal{R}\!\left(x,q,\hat{s};\,\mathcal{E}[\hat{s}]\right).
    \label{eq:decoupled}
\end{equation}
Both components are built on the same VLM, which the Perceiver equips with perception models and the Reasoner with the engine. The perceived facts are thereby exposed in $\hat{s}$ for inspection and evaluation, and the reasoning is grounded: the consequences of spatial operations are computed by the engine from $\hat{s}$ rather than generated by the VLM. The next two sections describe them.

\vspace{-.05in}
\subsection{Perceiver: Estimating the World State}
\label{sec:method:perceiver}
\label{sec:method:estimation}
\label{sec:method:instantiation}
\vspace{-.05in}

The Perceiver estimates $\hat{s}$ from $(x,q)$ (Figure~\ref{fig:architecture}, middle). The query determines which entities and properties the state must contain and may state some of the facts, whereas the images are the source of the observations and the camera poses. A text channel and a visual channel process the two inputs, for one image or several alike, and compilation turns their outputs into the layout.

\textbf{Text channel.} A VLM parses $q$ into four kinds of information. The \emph{entities} $\mathcal{O}$ are the objects and viewers the question involves. The \emph{relations} are facts that $q$ states about pairs of entities, such as one being to the left of another. The \emph{viewpoints} are facts that $q$ states about the cameras, such as the camera turning left between two images. The \emph{query} is the spatial question to be answered together with the types of information it requires, such as direction, distance, facing, or camera motion. Relations and viewpoints, when stated, form $\mathcal{C}$ and constrain the layout and the camera poses; the required information types later decide whether $\hat{s}$ can support the answer (Section~\ref{sec:method:reasoner}).

\textbf{Visual channel.} The visual channel estimates $\Omega_i$ for each image and the relative poses of the cameras when there are multiple images. It applies specialized models in four steps: \emph{detection} localizes each entity in $\mathcal{O}$ in the image, which with the camera intrinsics gives its direction; \emph{depth estimation} gives its distance; \emph{orientation estimation} gives its facing; and, for multiple images, \emph{camera pose estimation} gives the rotation and translation between cameras. The VLM also describes each image qualitatively, e.g., which entities are visible and how pairs are ordered, which covers entities the models fail to measure. If no observation relates a pair of entities the answer depends on, the Perceiver queries the VLM about this pair directly, so that every relation the question hinges on has at least one source.

\textbf{Compilation.} Compilation turns the observations and constraints into the layout. The shared frame is defined by the camera that took the first image $I_1$: this camera is placed at the origin and its viewing direction serves as the reference heading, so $(\mathbf{c}_1,\phi_1)=(\mathbf{0},0)$. The poses of the other cameras come from up to four sources: the viewpoints stated in $q$, the camera pose estimator, the entities seen in common with an earlier image, and the camera motion described by the VLM. We use the sources in a fixed order of priority, taking the first one that is available. For the heading, the order is stated viewpoints, common entities, pose estimator, VLM description; for the position, it is stated viewpoints, pose estimator, VLM description. The two components are chosen separately, and a component that no source provides remains unknown. Each observation is then projected into the shared frame. For an entity $j$ visible in image $I_i$, let $\mathbf{r}_{ij}$ denote its planar displacement in the camera frame and $\alpha_{ij}$ its facing relative to the camera's forward direction. The entity is placed at
\begin{equation}
    \mathbf{p}_{ij}=\mathbf{c}_i+\mathbf{R}(\phi_i)\,\mathbf{r}_{ij},
    \qquad
    \theta_{ij}=(\phi_i+\alpha_{ij})\bmod 2\pi,
    \label{eq:shared-frame-transform}
\end{equation}
where $\mathbf{R}(\phi_i)$ is the rotation by $\phi_i$. An entity seen in several images receives several candidates; its position $\hat{\mathbf{p}}_j$ is the candidate with the highest confidence, which accounts for the measurement and the camera pose, and its heading is a vote over all candidates. The candidates respect the measurements but may violate a stated relation or an observed pairwise order because of projection errors. We therefore search for the layout that best satisfies the relations while staying close to the candidates,
\begin{equation}
    \mathcal{L}^{*}=\arg\max_{\mathcal{L}}\ \sum_{k} w_k\,\mathds{1}\!\left[\mathcal{L}\models r_k\right]
    -\lambda\sum_{j}\big\|\mathbf{p}_j-\hat{\mathbf{p}}_j\big\|,
    \label{eq:layout}
\end{equation}
where $r_k$ ranges over the pairwise relations in $\mathcal{C}$ and in the observations, $\mathcal{L}\models r_k$ denotes that the layout satisfies $r_k$, and $w_k$ reflects the reliability of the source of $r_k$. The relations in $\mathcal{C}$ are hard constraints, the second term keeps the layout near the measurements, and the maximization is carried out by a deterministic local search that moves one entity at a time over the grid (Appendix~\ref{app:perceiver}).

\vspace{-.05in}
\subsection{Reasoner: Grounded Reasoning with a Physics Engine}
\label{sec:method:reasoner}
\label{sec:method:engine}
\label{sec:method:interaction}
\vspace{-.05in}

The Reasoner instantiates $\hat{s}$ in the physics engine and answers $q$ by querying it (Figure~\ref{fig:architecture}, right).

\begin{figure}[t]
\centering
\includegraphics[width=\textwidth]{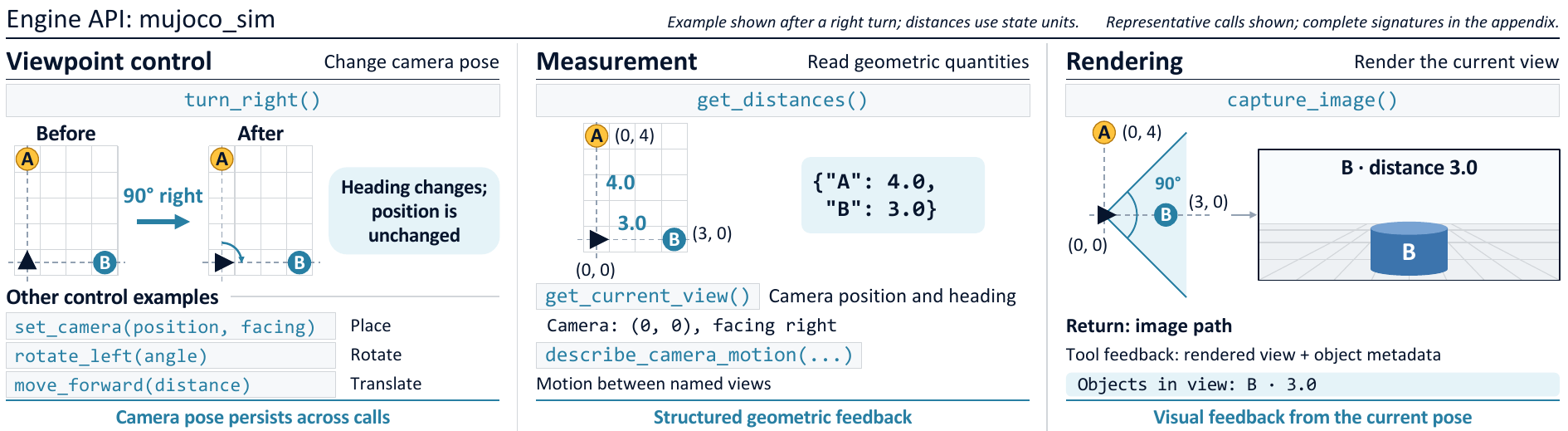}
\caption{\textbf{Engine interface exposed to the Reasoner.} Representative \texttt{mujoco\_sim} calls control the camera pose, return geometric measurements, and render views of the same scene. The pose persists across calls. Appendix~\ref{app:mujoco-interface} gives the full signatures, argument defaults, and return values.}
\vspace{-.15in}
\label{fig:api}
\end{figure}

\textbf{Physics engine as a tool.} A MuJoCo~\citep{todorov2012mujoco} scene is built from a scene description that gives each body a shape, either a mesh asset or a geometric primitive, and a pose. Generating a faithful mesh for each entity from the images is beyond what current models do reliably, so each entity is represented by a \emph{proxy} instead: a cylinder of fixed size at $(\mathbf{p}_j,\theta_j)$, with a heading marker and the entity's name as its label. Each input image corresponds to a camera at $(\mathbf{c}_i,\phi_i)$. The proxies carry exactly the information in the layout, on which the spatial questions depend. The resulting scene is $\mathcal{E}[\hat{s}]$. The Reasoner operates on it by writing code against a tool interface (Figure~\ref{fig:api}; full signatures in Listing~\ref{lst:sandbox}), and the engine returns printed results and rendered images. The operations form three groups:
\begin{itemize}[leftmargin=*,itemsep=1pt,topsep=2pt,parsep=0pt]
\vspace{-3pt}
\item \textit{Viewpoint control} places the camera at a given pose, such as the pose of an input image or a position described in the question, or moves and rotates it by given distances and angles. The camera keeps its pose between calls, so each operation starts from where the previous one left the camera, and a multi-step movement is executed step by step, one call at a time.
\item \textit{Measurement} returns the current camera pose, the distance from the camera to each entity, or the camera motion between two views, i.e., the translation and rotation from one to the other.
\item \textit{Rendering} returns the current view, with proxies labeled by name and distance.
\vspace{-3pt}
\end{itemize}
Every result is computed from $\hat{s}$, so the feedback is deterministic and consistent with it.

\textbf{Interactive reasoning.} With the scene in place, the Reasoner answers $q$ in a loop with the engine. The context $h_1$ initially holds $x$, $q$, $\hat{s}$, and a description of each operation in the interface, giving its purpose, arguments, and return value. At step $t$, the Reasoner reads the context, issues an operation $a_t$, and receives the feedback $f_t$, i.e., the printed results and, if requested, the rendered view, which is appended to the context together with the operation:
\begin{equation}
    a_t=\mathcal{R}(h_t),\qquad
    f_t=\mathcal{E}[\hat{s}](a_t),\qquad
    h_{t+1}=h_t\oplus(a_t,f_t).
    \label{eq:reasoner-feedback}
\end{equation}
In this loop, the Reasoner chooses operations whose results answer the question. For example, to decide whether moving forward brings the observer closer to an entity, it measures the distance before and after the movement; to find what is on the observer's left after a turn, it performs the turn, faces left, and renders the view. Each result confirms or revises its current judgment and determines the next operation, and the loop ends when the Reasoner answers or exhausts a budget of $T_{\max}$ steps; an answer is accepted only after at least one engine call. A routing step decides beforehand whether the engine is used: the loop is entered only when the information types that $q$ requires are among those the state describes; otherwise, the VLM answers directly from the images (Appendix~\ref{app:mujoco-interface}).

\newlength{\savedtextfloatsep}\setlength{\savedtextfloatsep}{\textfloatsep}
\newlength{\savedfloatsep}\setlength{\savedfloatsep}{\floatsep}
\setlength{\textfloatsep}{12pt plus 2pt}
\setlength{\floatsep}{11pt plus 2pt}
\vspace{-.15in}
\section{Experiments}
\label{sec:experiments}
\vspace{-.1in}

\newcommand{\expresult}{\textcolor{red}{XX}}
\newcommand{\expvalue}{\textcolor{red}{XX}}

\subsection{Experimental Setup}
\label{subsec:exp-setting}
\vspace{-.05in}

\textbf{Benchmarks.}
We evaluate on four spatial reasoning benchmarks: MindCube-Tiny~\citep{yin2025spatial}, SAT~\citep{ray2024sat}, MMSI-Bench~\citep{yang2026mmsi}, and SpatialRGPT-Bench~\citep{cheng2024spatialrgpt}.
MindCube-Tiny (MindCube hereafter) contains $1{,}050$ multiple-choice questions over a few real-image views, grouped by camera motion into \textit{Rotation}, \textit{Among}, and \textit{Around}.
For SAT, we sample $500$ questions from its synthetic split, $100$ from each of five categories.
MMSI-Bench contains $1{,}000$ multiple-choice questions over multiple real-scene images.
The quantitative split of SpatialRGPT-Bench contains $749$ questions that require numerical estimates.
We report answer accuracy on the three multiple-choice benchmarks; on SpatialRGPT-Bench, a prediction is correct if it is within the error tolerance.
Appendix~\ref{app:exp-data} provides further details.

\begin{table}[!t]
\centering
\captionsetup{skip=3pt}
\caption{MindCube accuracy (\%) in the GT-map setting. \textit{Base} (w/o map) answers from the images and question; \textit{Base} (w map) and Spatial Harness (w map) receive the same GT map. Parentheses give the gain over \textit{Base} (w map). Bold: best overall accuracy for each model; underline: second best.}
\label{tab:mainresults-mindcube-map}
\fontsize{8}{10.5}\selectfont
\setlength{\tabcolsep}{2.5pt}
\begin{tabular*}{\textwidth}{@{\extracolsep{\fill}}llcccc@{}}
\toprule
\textbf{Model} & \textbf{Method} & \textbf{Rotation} & \textbf{Among} & \textbf{Around} & \textbf{Overall} \\
\midrule
\multirow{3}{*}{Qwen2.5-VL-7B-Instruct} & Base (w/o map) & 35.83 & 33.39 & 40.67 & \underline{35.59} \\
 & Base (w map) & 35.50 & 27.56 & 32.93 & 30.35 \\
 & Spatial Harness (w map) & 34.33 & 58.17 & 70.27 & \textbf{56.51 (+26.16)} \\
\midrule
\multirow{3}{*}{Qwen2.5-VL-32B-Instruct} & Base (w/o map) & 38.83 & 38.17 & 46.00 & 40.16 \\
 & Base (w map) & 40.83 & 40.11 & 65.73 & \underline{46.35} \\
 & Spatial Harness (w map) & 69.00 & 88.39 & 84.53 & \textbf{83.78 (+37.43)} \\
\midrule
\multirow{3}{*}{Qwen2.5-VL-72B-Instruct} & Base (w/o map) & 40.83 & 41.06 & 42.93 & \underline{41.46} \\
 & Base (w map) & 35.67 & 45.28 & 36.93 & \underline{41.46} \\
 & Spatial Harness (w map) & 79.50 & 92.56 & 68.00 & \textbf{84.22 (+42.76)} \\
\midrule
\multirow{3}{*}{Qwen3.5-27B} & Base (w/o map) & 77.17 & 41.72 & 59.47 & 52.70 \\
 & Base (w map) & 83.83 & 60.94 & 88.27 & \underline{71.81} \\
 & Spatial Harness (w map) & 96.83 & 98.06 & 99.87 & \textbf{98.25 (+26.44)} \\
\midrule
\multirow{3}{*}{Kimi K2.5} & Base (w/o map) & 69.67 & 42.78 & 54.40 & 50.67 \\
 & Base (w map) & 96.00 & 75.83 & 82.53 & \underline{81.27} \\
 & Spatial Harness (w map) & 99.67 & 95.33 & 86.67 & \textbf{94.10 (+12.83)} \\
\midrule
\multirow{3}{*}{GPT-5.4-mini} & Base (w/o map) & 28.83 & 42.72 & 60.27 & 44.25 \\
 & Base (w map) & 27.83 & 42.61 & 63.60 & \underline{44.79} \\
 & Spatial Harness (w map) & 71.83 & 92.44 & 68.67 & \textbf{82.86 (+38.07)} \\
\midrule
\multirow{3}{*}{GLM-5V-Turbo} & Base (w/o map) & 74.67 & 56.44 & 73.87 & 64.06 \\
 & Base (w map) & 77.83 & 56.83 & 84.67 & \underline{67.46} \\
 & Spatial Harness (w map) & 90.17 & 97.22 & 71.20 & \textbf{89.68 (+22.22)} \\
\bottomrule
\end{tabular*}
\vspace{-.1in}
\end{table}

\vspace{-.1in}
\paragraph{Baselines.}
We compare with three baselines: \textit{Base}, \textit{MindJourney}~\citep{yang2025mindjourney}, and \textit{pySpatial}~\citep{luo2026pySpatial}.
\textit{Base} prompts the VLM to answer without external tools.
\textit{MindJourney} uses a video world model to generate views along camera trajectories proposed by the VLM, which then answers from the generated views.
\textit{pySpatial} lets the VLM write Python programs that call 3D reconstruction, camera-pose, and novel-view tools.
Appendix~\ref{app:gca} further compares with \textit{GCA}~\citep{chen2025geometrically}.

\vspace{-.1in}
\paragraph{Evaluation settings.}
We evaluate in two settings.
The \textit{GT-map} setting uses the annotated cognitive map that MindCube provides for each question, which gives the ground-truth scene layout; we call it the GT map.
It marks the positions and facing directions of the relevant objects and cameras on a $10\times10$ bird's-eye-view grid.
\textit{Base} reads the map as text in its prompt, whereas Spatial Harness instantiates the map in the engine and the Reasoner queries it.
Because both methods start from the same scene information, this setting compares how they reason with it, separately from errors in state estimation.
In the \textit{end-to-end} setting, methods receive only the images and question.
Spatial Harness builds its world state with the Perceiver, and the baselines use their own tools.

\vspace{-.1in}
\paragraph{Vision-language models.}
All methods in a comparison use the same VLM.
In the GT-map setting, we evaluate seven open- and closed-weight models: Qwen2.5-VL-7B/32B/72B-Instruct, Qwen3.5-27B, Kimi K2.5, GPT-5.4-mini, and GLM-5V-Turbo~\citep{hong2026glm}.
In the end-to-end setting, we evaluate Qwen2.5-VL-72B-Instruct, Qwen3.5-27B, Kimi K2.5, and GPT-5.4-mini on MindCube.
Because of the evaluation cost, we use only the two Qwen models on the other three benchmarks.

\vspace{-.1in}
\paragraph{Spatial Harness setup.}
In the end-to-end setting, the evaluated VLM serves as both the VLM in the Perceiver and the Reasoner.
The state is instantiated in MuJoCo~\citep{todorov2012mujoco}, and the Reasoner takes at most $T_{\max}=5$ turns.
Appendix~\ref{app:implementation-details} gives further implementation details.

\subsection{Main Results}
\label{subsec:exp-main}

\begin{table}[!t]
\centering
\captionsetup{skip=3pt}
\caption{End-to-end accuracy (\%) on MindCube. Spatial Harness reasons over the world state estimated by its Perceiver. Bold: best overall accuracy for each model; underline: second best.}
\label{tab:mainresults-mindcube}
\fontsize{8}{10.5}\selectfont
\setlength{\tabcolsep}{4pt}
\begin{tabular*}{\textwidth}{@{\extracolsep{\fill}}llcccc@{}}
\toprule
\textbf{Model} & \textbf{Method} & \textbf{Rotation} & \textbf{Among} & \textbf{Around} & \textbf{Overall} \\
\midrule
\multirow{4}{*}{Qwen2.5-VL-72B-Instruct} & Base & 40.83 & 41.06 & 42.93 & 41.46 \\
 & MindJourney & 42.00 & 40.50 & 49.20 & 42.86 \\
 & pySpatial & 30.50 & 55.50 & 53.60 & \underline{50.29} \\
 & Spatial Harness & 49.50 & 58.00 & 53.60 & \textbf{55.33} \\
\midrule
\multirow{4}{*}{Qwen3.5-27B} & Base & 77.17 & 41.72 & 59.47 & 52.70 \\
 & MindJourney & 82.00 & 42.00 & 65.60 & 55.24 \\
 & pySpatial & 61.50 & 59.00 & 76.40 & \underline{63.62} \\
 & Spatial Harness & 81.50 & 68.00 & 70.40 & \textbf{71.14} \\
\midrule
\multirow{4}{*}{Kimi K2.5}& Base & 69.67 & 42.78 & 54.40 & 50.67 \\
 & MindJourney & 68.00 & 42.50 & 53.60 & 50.00 \\
 & pySpatial & 62.50 & 65.17 & 75.60 & \underline{67.14} \\
 & Spatial Harness & 85.50 & 68.17 & 73.20 & \textbf{72.67} \\
\midrule
\multirow{4}{*}{GPT-5.4-mini}& Base & 28.83 & 42.72 & 60.27 & 44.25 \\
 & MindJourney & 30.00 & 50.33 & 59.60 & 48.67 \\
 & pySpatial & 38.50 & 56.33 & 69.60 & \underline{56.10} \\
 & Spatial Harness & 49.50 & 59.83 & 67.20 & \textbf{59.62} \\
\bottomrule
\end{tabular*}
\vspace{-.1in}
\end{table}

\paragraph{Reasoning with GT maps.}
Figure~\ref{fig:mindcube-gt-comparison} summarizes overall accuracy with and without GT maps.
Table~\ref{tab:mainresults-mindcube-map} reports MindCube accuracy in the GT-map setting.
Providing the map for direct reasoning does not always help and can even hurt: with the map, \textit{Base} accuracy drops from $35.59\%$ to $30.35\%$ with Qwen2.5-VL-7B.
Some recent models make better use of the map, which raises \textit{Base} accuracy by $19.11$ points with Qwen3.5-27B and $30.60$ points with Kimi K2.5.
Spatial Harness instead grounds each reasoning step in engine feedback.
With the same map, it reaches $84.22\%$ with Qwen2.5-VL-72B and $98.25\%$ with Qwen3.5-27B, $42.76$ and $26.44$ points above \textit{Base} with the map.
Grounded feedback thus improves reasoning whether or not the textual map helps \textit{Base}.

\vspace{-.1in}
\paragraph{End-to-end performance.}
Table~\ref{tab:mainresults-mindcube} reports end-to-end results on MindCube, where Spatial Harness reasons over the world state estimated by its Perceiver.
It achieves the highest overall accuracy with all four VLMs, exceeding the strongest baseline, \textit{pySpatial}, by $5.40$ points on average.
With Qwen3.5-27B, it reaches $71.14\%$, $18.44$ points above \textit{Base} and $7.52$ points above \textit{pySpatial}.
Spatial Harness differs from the baselines in where its feedback comes from: the world state takes in both the images and the question, and every measurement and rendered view is computed from that state.
\textit{MindJourney} instead imagines new viewpoints with a video world model, whose generated views are not constrained by an explicit scene, and it stays within $4.42$ points of \textit{Base}.
\textit{pySpatial} reasons over a scene reconstructed from pixels, which cannot capture spatial information given only in the question, such as the camera motion between views, and it falls below \textit{Base} on some question categories.

Table~\ref{tab:mainresults-crossbenchmark} reports end-to-end results on the other three benchmarks.
Averaged over the three benchmarks, Spatial Harness achieves the highest accuracy with both models, $15.26$ and $14.20$ points above the strongest baseline.
On SAT, most questions provide a single image, and the VLM already answers many of them well.
Reconstructing a 3D scene from a single image is difficult, and \textit{pySpatial}, which reasons over such reconstructions, falls $13.20$ and $13.80$ points below \textit{Base}.
Spatial Harness achieves the best accuracy with both models ($71.60\%$ and $83.20\%$).
On MMSI-Bench, it improves over \textit{Base} with Qwen3.5-27B from $33.40\%$ to $39.50\%$.
With Qwen2.5-VL-72B, it remains at the \textit{Base} level ($29.00\%$ vs.\ $29.60\%$), because the model often misreads the coordinates in the world state and does not act on the engine's feedback in its reasoning.
On SpatialRGPT-Bench, it reaches $84.11\%$ and $83.31\%$, about $43$ and $35$ points above the strongest baseline, as it obtains distances and object sizes from metric perception tools and engine measurements rather than appearance.

\begin{table}[!t]
\centering
\captionsetup{skip=3pt}
\caption{End-to-end results (\%) on SAT, MMSI-Bench, and SpatialRGPT-Bench. SAT and MMSI-Bench report answer accuracy, and SpatialRGPT-Bench reports the fraction of predictions within its error tolerance. Avg is the mean over the three benchmarks. Per-category results are in Tables~\ref{tab:mainresults-sat-detail}--\ref{tab:mainresults-spatialrgpt}. Bold: best result for each model and column; underline: second best.}
\label{tab:mainresults-crossbenchmark}
\fontsize{8}{10.5}\selectfont
\setlength{\tabcolsep}{3pt}
\begin{tabular*}{\textwidth}{@{\extracolsep{\fill}}llcccc@{}}
\toprule
\textbf{Model} & \textbf{Method} & \textbf{SAT} & \textbf{MMSI-Bench} & \textbf{SpatialRGPT-Bench} & \textbf{Avg} \\
\midrule
\multirow{4}{*}{Qwen2.5-VL-72B-Instruct} & Base & \underline{68.60} & 29.60 & \underline{40.72} & \underline{46.31} \\
 & MindJourney & \underline{68.60} & \underline{30.60} & 39.12 & 46.11 \\
 & pySpatial & 55.40 & \textbf{33.20} & 31.64 & 40.08 \\
 & Spatial Harness & \textbf{71.60} & 29.00 & \textbf{84.11} & \textbf{61.57} \\
\midrule
\multirow{4}{*}{Qwen3.5-27B} & Base & \underline{81.80} & 33.40 & \underline{48.20} & \underline{54.47} \\
 & MindJourney & 78.40 & 33.00 & 45.66 & 52.35 \\
 & pySpatial & 68.00 & \underline{36.70} & 39.79 & 48.16 \\
 & Spatial Harness & \textbf{83.20} & \textbf{39.50} & \textbf{83.31} & \textbf{68.67} \\
\bottomrule
\end{tabular*}
\vspace{-.1in}
\end{table}

\subsection{Ablation Studies}
\label{subsec:exp-ablation}

\paragraph{World state construction.}
To examine how the world state should be constructed, we fix the Reasoner and vary the source of the map on MindCube with Qwen3.5-27B (Table~\ref{tab:ablation-mindcube}).
In the VLM-generated condition, the VLM produces the map directly from the images and question without the Perceiver.
Accuracy drops from $71.14\%$ to $41.33\%$: the VLM alone cannot reliably estimate the state, and the Perceiver's structured estimation is needed.
We then ablate the two perception channels.
Without the visual channel, the VLM estimates the geometric fields in place of the specialized visual models, and accuracy drops to $52.95\%$.
Without the text channel, the VLM identifies the relevant entities from the images instead of extracting them from the question, and accuracy drops to $68.38\%$.
Both channels are therefore needed, and the visual channel contributes more.

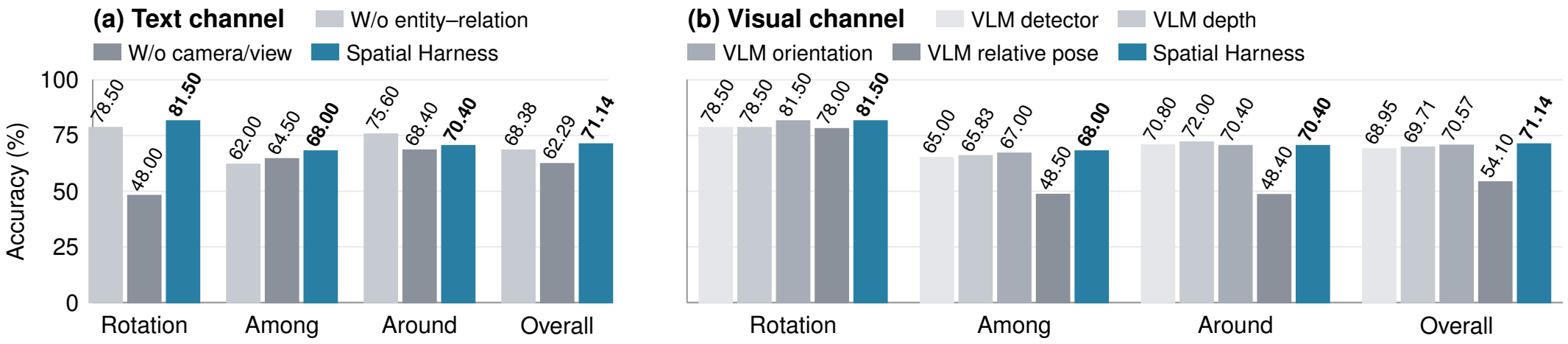
\begin{figure}[!t]
\centering
\captionsetup{skip=3pt}
\makebox[\textwidth][c]{%
\begin{tikzpicture}
\begin{axis}[
  name=chanA, chartbase, ybar=1.5pt, chartvalue, bar width=8.4pt, width=132pt, height=2cm, ymin=0, ymax=100, ytick={0,25,50,75,100},
  ylabel={Accuracy (\%)}, ylabel style={yshift=-4pt}, symbolic x coords={Rotation,Among,Around,Overall}, xtick=data, enlarge x limits=0.125,
]
\addplot+[chartsolid=chartGrayB] coordinates {(Rotation,78.50) (Among,62.00) (Around,75.60) (Overall,68.38)};
\addplot+[chartsolid=chartGrayD] coordinates {(Rotation,48.00) (Among,64.50) (Around,68.40) (Overall,62.29)};
\addplot+[chartsh, chartvaluesh] coordinates {(Rotation,81.50) (Among,68.00) (Around,70.40) (Overall,71.14)};
\end{axis}
\begin{axis}[
  name=chanB, at={(chanA.south east)}, anchor=south west, xshift=20pt, chartbase, ybar=1.5pt, chartvalue, bar width=8.4pt, width=224pt, height=2cm, ymin=0, ymax=100, ytick={0,25,50,75,100}, yticklabels={},
  symbolic x coords={Rotation,Among,Around,Overall}, xtick=data, enlarge x limits=0.16,
]
\addplot+[chartsolid=chartGrayA] coordinates {(Rotation,78.50) (Among,65.00) (Around,70.80) (Overall,68.95)};
\addplot+[chartsolid=chartGrayB] coordinates {(Rotation,78.50) (Among,65.83) (Around,72.00) (Overall,69.71)};
\addplot+[chartsolid=chartGrayC] coordinates {(Rotation,81.50) (Among,67.00) (Around,70.40) (Overall,70.57)};
\addplot+[chartsolid=chartGrayD] coordinates {(Rotation,78.00) (Among,48.50) (Around,48.40) (Overall,54.10)};
\addplot+[chartsh, chartvaluesh] coordinates {(Rotation,81.50) (Among,68.00) (Around,70.40) (Overall,71.14)};
\end{axis}
\node[font=\sffamily\fontsize{5.5}{6.5}\selectfont, inner sep=0pt, anchor=base west] at ($(chanA.north west)+(0pt,13.5pt)$) {{\bfseries\fontsize{6.5}{7.5}\selectfont (a) Text channel}\hspace{8pt}\chartkey{chartGrayB}{W/o entity--relation}};
\node[font=\sffamily\fontsize{5.5}{6.5}\selectfont, inner sep=0pt, anchor=base west] at ($(chanA.north west)+(0pt,5pt)$) {\chartkey{chartGrayD}{W/o camera/view}\hspace{5pt}\chartkey{chartSH}{Spatial Harness}};
\node[font=\sffamily\fontsize{5.5}{6.5}\selectfont, inner sep=0pt, anchor=base west] at ($(chanB.north west)+(0pt,13.5pt)$) {{\bfseries\fontsize{6.5}{7.5}\selectfont (b) Visual channel}\hspace{8pt}\chartkey{chartGrayA}{VLM detector}\hspace{5pt}\chartkey{chartGrayB}{VLM depth}};
\node[font=\sffamily\fontsize{5.5}{6.5}\selectfont, inner sep=0pt, anchor=base west] at ($(chanB.north west)+(0pt,5pt)$) {\chartkey{chartGrayC}{VLM orientation}\hspace{5pt}\chartkey{chartGrayD}{VLM relative pose}\hspace{5pt}\chartkey{chartSH}{Spatial Harness}};
\end{tikzpicture}}
\vspace{-.17in}
\caption{Channel ablations on MindCube with Qwen3.5-27B. (a) Each condition removes one module from the text channel; the entity--relation module extracts entities and relations jointly. (b) Each condition replaces one specialized model with the VLM, which produces the same output fields.}
\label{fig:channel-ablation}
\vspace{-.15in}
\end{figure}

We also ablate the design within each channel (Figure~\ref{fig:channel-ablation}).
In the text channel, dropping the camera and view constraints lowers accuracy to $62.29\%$, mainly on \textit{Rotation} ($81.50\%$ to $48.00\%$), whose questions state the camera rotation.
In the visual channel, replacing the relative-pose estimator costs the most ($54.10\%$), mainly on \textit{Among} and \textit{Around}, where the camera changes position between views; replacing the orientation estimator costs the least ($70.57\%$) because only $36$ of the $1{,}050$ questions depend on an object's facing.
Every variant lowers overall accuracy, which supports the structured extraction and the use of specialized models; Appendix~\ref{app:exp-ablations} describes each intervention.

\begin{table}[!t]
\centering
\captionsetup{skip=3pt}
\caption{Ablation on world state construction on MindCube with Qwen3.5-27B (accuracy in \%). The Reasoner is fixed, and only the source of the map changes; the GT map serves as a reference.}
\label{tab:ablation-mindcube}
\label{tab:ablation-channels}
\fontsize{8}{10.5}\selectfont
\setlength{\tabcolsep}{4pt}
\begin{tabular*}{\textwidth}{@{\extracolsep{\fill}}lcccc@{}}
\toprule
\textbf{Map source} & \textbf{Rotation} & \textbf{Among} & \textbf{Around} & \textbf{Overall} \\
\midrule
VLM-generated map & 54.00 & 28.83 & 61.20 & 41.33 \\
Perceiver without text channel & 71.00 & 63.00 & 79.20 & 68.38 \\
Perceiver without visual channel & 72.50 & 49.67 & 45.20 & 52.95 \\
Full Perceiver (Spatial Harness) & 81.50 & 68.00 & 70.40 & 71.14 \\
GT map (reference) & 96.83 & 98.06 & 99.87 & 98.25 \\
\bottomrule
\end{tabular*}
\vspace{-.1in}
\end{table}

\vspace{-.1in}
\paragraph{Grounded feedback.}
To examine where the gains of the Reasoner come from, we fix the GT map, which excludes perception errors, and vary only the feedback available to the Reasoner on MindCube with Qwen3.5-27B (Table~\ref{tab:ablation-backend}).
\textit{Base} answers directly from the map.
In multi-turn self-review, the VLM reviews its initial answer in further turns without receiving new scene information.
Self-review raises accuracy only from $71.81\%$ to $77.81\%$, showing that the gains do not come from additional turns.
Either type of engine feedback alone raises accuracy above $90\%$, and combining both achieves the highest accuracy of $98.25\%$, so the two types of feedback are complementary.
Appendix~\ref{app:feedback-conditions} describes each condition, and Appendix~\ref{app:exp-cases} gives example trajectories.

\begin{table}[!t]
\centering
\captionsetup{skip=3pt}
\caption{Ablation on grounded feedback on MindCube with Qwen3.5-27B (accuracy in \%). All conditions reason over the same GT map and differ only in the feedback available to the Reasoner.}
\label{tab:ablation-backend}
\fontsize{8}{10.5}\selectfont
\setlength{\tabcolsep}{4pt}
\begin{tabular*}{\textwidth}{@{\extracolsep{\fill}}lcccc@{}}
\toprule
\textbf{Feedback condition} & \textbf{Rotation} & \textbf{Among} & \textbf{Around} & \textbf{Overall} \\
\midrule
Base & 83.83 & 60.94 & 88.27 & 71.81 \\
Multi-turn self-review & 96.00 & 66.67 & 90.00 & 77.81 \\
Numerical feedback only & 93.00 & 95.00 & 99.20 & \underline{95.62} \\
Rendered feedback only & 95.00 & 92.33 & 96.00 & 93.71 \\
Spatial Harness & 96.83 & 98.06 & 99.87 & \textbf{98.25} \\
\bottomrule
\end{tabular*}
\end{table}

\vspace{-.05in}
\section{Discussion}
\label{sec:discuss}
\vspace{-.05in}

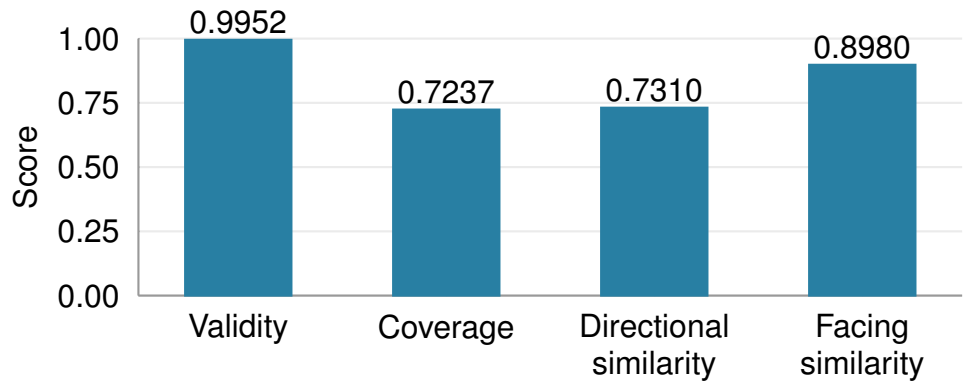
\begin{wrapfigure}{r}{0.52\textwidth}
\vspace{-.15in}
\centering
\captionsetup{skip=3pt}
\begin{tikzpicture}
\begin{axis}[
  chartbase, width=0.39\textwidth, height=1.7cm, bar width=20pt, ymin=0, ymax=1, ytick={0,0.25,0.5,0.75,1},
  yticklabel style={/pgf/number format/fixed, /pgf/number format/fixed zerofill, /pgf/number format/precision=2},
  ylabel={Score}, ylabel style={yshift=-3pt}, symbolic x coords={Validity,Coverage,Directional,Facing}, xtick=data, enlarge x limits=0.16,
  xticklabels={Validity, Coverage, {Directional\\similarity}, {Facing\\similarity}}, xticklabel style={align=center},
  nodes near coords, every node near coord/.append style={font=\sffamily\fontsize{6}{7}\selectfont, text=black, inner sep=1pt, /pgf/number format/fixed, /pgf/number format/fixed zerofill, /pgf/number format/precision=4},
]
\addplot+[chartsh] coordinates {(Validity,0.9952) (Coverage,0.7237) (Directional,0.7310) (Facing,0.8980)};
\end{axis}
\end{tikzpicture}
\vspace{-.05in}
\caption{Quality of the world states estimated by the Perceiver with Qwen3.5-27B on MindCube. Maps are scored against the GT maps with the MindCube cognitive-map metrics after name alignment (Appendix~\ref{app:exp-state-metrics}); scores lie in $[0,1]$ (higher is better).}
\label{fig:state-quality}
\vspace{-.1in}
\end{wrapfigure}

\paragraph{World State Quality.}
Since the Reasoner works on the world state estimated by the Perceiver, we examine how closely this state matches the scene.
We compare the world states that the Perceiver estimates with \mbox{Qwen3.5-27B} against the GT maps of all $1{,}050$ MindCube questions.
Following MindCube~\citep{yin2025spatial}, we score each map with its cognitive-map metrics (Appendix~\ref{app:exp-state-metrics}): \emph{validity}, \emph{coverage}, and \emph{directional} and \emph{facing similarity}.
Nearly all estimated maps are valid ($99.52\%$), and they recover $72\%$ of the annotated entities after name alignment (Figure~\ref{fig:state-quality}).
With these maps, Spatial Harness already achieves the best end-to-end accuracy, and its higher accuracy with GT maps suggests that more accurate maps would raise it further; Appendix~\ref{app:state-noise-results} measures how accuracy degrades when single fields of the GT map are corrupted.
The grounded feedback of the engine could also be used to check the world state itself.
The engine can render it from the estimated pose of each input view.
A rendering that disagrees with the corresponding input image would locate an error for the Perceiver to correct.

\vspace{-.1in}
\paragraph{Limitations.}
Our world state describes a static scene at a single moment, which suffices for the spatial questions in our benchmarks but cannot represent change over time, such as objects moving in a video. A time dimension is the natural next step: recording the poses of entities and the camera at each frame would let the engine instantiate the scene at any moment and measure how it changes between frames, which would support questions about motion and the order of events. We plan to support this temporal dimension in future work, together with the video benchmarks that require it.

\vspace{-.05in}
\section{Conclusion}
\vspace{-.05in}
Answering a spatial question involves perception, which establishes the scene facts, and reasoning, which can be computed exactly once those facts are explicit.
We therefore propose Spatial Harness, which separates the two steps: a Perceiver estimates the scene facts from the images and the question as an explicit world state, and a Reasoner answers by operating on this state in a physics engine, so that every measurement and rendered view is computed from the state rather than imagined.
Experiments on four spatial reasoning benchmarks show that Spatial Harness outperforms strong baselines, and ablations attribute the gains to both the explicit state and the grounded feedback.
We hope this work encourages the use of physics engines as a general substrate for grounding multimodal reasoning about the physical world, where perception and reasoning can be checked separately.

\clearpage
\setlength{\textfloatsep}{\savedtextfloatsep}
\setlength{\floatsep}{\savedfloatsep}
\begin{ack}
CTC, ZKZ, and BH were supported by NSFC Major Research Plan No. 92570109, RGC Research Fellow Scheme No. RFS2627-2S02, RGC Young Collaborative Research Grant No. C2005-24Y, RGC General Research Fund No. 12202026, RGC General Research Fund No. 12200725, NSFC General Program No. 62376235, ByteDance Faculty Research Award, Guangdong and Hong Kong Universities ``1+1+1'' Joint Research Collaboration Scheme, HKBU Faculty Niche Research Areas No. RC-FNRA-IG/25-26/SCI/05, and HKBU CSD Departmental Incentive Scheme. The authors also thank Jingkun Zhang, Zhida Zhou, Shiyang Liu, and Jiaxuan Wang for their assistance with the experiments. The authors declare no competing interests.
\end{ack}

\bibliographystyle{plainnat}
\bibliography{ref}

\clearpage
\appendix
\part{Appendix}
\etocdepthtag.toc{mtappendix}
\etocsettagdepth{mtchapter}{none}
\etocsettagdepth{mtappendix}{subsection}
\renewcommand{\contentsname}{}
\tableofcontents
\newpage

\section{Broader Impacts}\label{app:impact}
Spatial Harness may benefit embodied agents by improving spatial reasoning through explicit world states and executable feedback. Making scene interpretations and computed evidence available for inspection could support the development of more reliable and transparent embodied systems. Since errors in scene estimation or reasoning can still affect downstream decisions, applications in physical environments should include appropriate validation.

\section{Use of Large Language Models}
\label{app:llm-usage}
Vision-language models are core components of Spatial Harness: in each experiment, the evaluated VLM serves as the VLM in the Perceiver and as the Reasoner (Section~\ref{sec:methodology}), the baselines use the same VLM, and Section~\ref{subsec:exp-setting} lists the models. An LLM judge scores free-form answers against the reference answers (Appendix~\ref{app:exp-evaluation}), and the name judge of the map-quality evaluation is an LLM as well (Appendix~\ref{app:exp-state-metrics}). Beyond these roles in the method and the evaluation, LLMs assisted only in preparing the manuscript, where they were used to polish the wording and check the grammar; they did not produce the research ideas, the experiments, or the results. The authors have reviewed the whole text and take full responsibility for its content, including its originality and factual accuracy. No LLM is an author of this paper.

\section{Implementation Details}
\label{app:implementation-details}

\subsection{Spatial Harness Algorithm}
\label{app:algorithm}

Algorithm~\ref{alg:visual-harness} summarizes world state construction and engine-grounded reasoning in Spatial Harness.
\begin{algorithm}[!htbp]
\caption{Spatial Harness}
\label{alg:visual-harness}
\small
\renewcommand{\algorithmiccomment}[1]{\hfill $\triangleright$~#1}
\begin{algorithmic}[1]
\REQUIRE Images $x$, query $q$; $\mathcal{P}$, $\mathcal{R}$, $\mathcal{E}$; turn budget $T_{\max}$
\ENSURE Predicted answer $\hat{y}$
\STATE \textbf{Perceiver $\mathcal{P}$} \COMMENT{Section~\ref{sec:method:perceiver}}
\STATE Parse $q$ into entities, relations, viewpoints, and the query \COMMENT{Text channel}
\STATE Estimate per-view observations and camera poses from $x$ \COMMENT{Visual channel}
\STATE Compile the observations into the world state $\hat{s}$
\STATE \textbf{Reasoner $\mathcal{R}$} \COMMENT{Section~\ref{sec:method:reasoner}}
\IF{$\hat{s}$ cannot support the answer}
    \RETURN the answer of the VLM from $(x,q)$
\ENDIF
\STATE Instantiate $\hat{s}$ in $\mathcal{E}$; $h_1 \leftarrow (x,q,\hat{s},\text{API descriptions})$
\FOR{$t=1,\ldots,T_{\max}$}
    \STATE $\hat{y}\ \text{or}\ a_t \leftarrow \mathcal{R}(h_t)$
    \IF{an answer $\hat{y}$ is returned}
        \IF{the engine has been called \textbf{or} $t=T_{\max}$}
            \RETURN $\hat{y}$
        \ELSE
            \STATE Ask the Reasoner to call the engine to obtain $h_{t+1}$
        \ENDIF
    \ELSE
        \STATE $f_t \leftarrow \mathcal{E}[\hat{s}](a_t)$ \COMMENT{Engine feedback}
        \STATE Append $(a_t,f_t)$ and the latest rendered view to obtain $h_{t+1}$
    \ENDIF
\ENDFOR
\STATE Request a final answer from $\mathcal{R}$ without further tool use
\RETURN this answer, or the answer of the VLM from $(x,q)$ if it is not valid
\end{algorithmic}
\end{algorithm}

\subsection{Perceiver}
\label{app:perceiver}

\paragraph{Perception models.}
The text channel and the qualitative image descriptions use the evaluated VLM. LocateAnything-3B~\citep{wang2026locateanything} detects each entity by name, and Qwen3-Reranker-0.6B~\citep{zhang2025qwen3embedding} binds detector labels to entity names when the two do not match exactly. MoGe-2~\citep{wang2025moge} estimates metric depth and camera intrinsics, and the depth of an entity is the median depth over the central half of its box. Orient Anything V2~\citep{wang2026orient} estimates the facing direction of an entity in the camera frame, which is kept when its front confidence is at least $0.5$. For multiple images, Reloc3r~\citep{wang2024reloc3r} estimates the relative rotation and the translation direction between the first image and each other image. The perception models are served on A100 GPUs.

\paragraph{Image descriptions and pair queries.}
For each image, the VLM reports whether each entity is visible, its coarse horizontal position (left, center, or right), its depth rank, its coarse facing, and the left--right and front--back order of entity pairs. A detected box that the VLM does not confirm as visible receives a low confidence. When the question names fewer than two entities, the VLM proposes landmarks in each image, which locate the cameras and are placed on the map but do not count as question entities. Pair queries cover at most eight entity pairs for which the image descriptions give no left--right or in-front-of order, starting with the pairs that involve the reference entity of the question. A pair visible in one image is queried in that image for its left--right and front--back order; a pair seen in different images but never together is queried across two images, and only its left--right order is kept.

\paragraph{Camera poses.}
Each camera pose has three components: two position coordinates and a heading. Each source supplies the components it can measure with a fixed confidence. For the heading, these are the text ($0.95$), landmark correspondences between images ($0.6$), Reloc3r ($0.5$), and a VLM estimate of the motion from the previous image ($0.4$, capped by the confidence of the previous pose); for the position, the text ($0.95$), Reloc3r ($0.7$), and the VLM estimate ($0.4$, capped likewise). The VLM estimate is requested for every image after the first, except an image whose pose the question states and whose components the text, landmark, and Reloc3r sources already cover. A landmark heading that reports no rotation is demoted below Reloc3r when Reloc3r measures a turn of at least $15^\circ$, and a landmark heading chained through an earlier image ranks below a direct one. Each component is taken separately from its most confident source, and a component that no source measures is marked as unknown, except that a camera motion stated in the text sets its omitted displacement components to zero, and a stated rotation in place keeps the previous position and its confidence in preference to visual estimates unless the text gives absolute coordinates.

\paragraph{Coordinate projection.}
\label{app:coordinate-projection}
The shared frame is anchored at the camera that took the first image. In Equation~\eqref{eq:shared-frame-transform}, the planar displacement of entity $j$ in camera $i$ is
\begin{equation}
    \mathbf{r}_{ij}=\frac{d_{ij}}{\cos\beta_{ij}}\,\mathbf{u}(\beta_{ij}),
    \label{eq:continuous-projection}
\end{equation}
where $d_{ij}$ is its depth along the optical axis, $\beta_{ij}$ its horizontal bearing from the image location and the intrinsics, and $\mathbf{u}(\theta)$ the unit vector at angle $\theta$ from the camera's forward direction. The factor $1/\cos\beta_{ij}$ converts optical-axis depth into the distance along the ray. Orient Anything V2 reports an azimuth $a_{ij}$ whose zero direction faces the camera, which we convert to the orientation $\alpha_{ij}=\pi-a_{ij}$ relative to the camera's forward direction.

\paragraph{Fusion and layout.}
Each observation of an entity is weighted by the product of its confidence (at least $0.05$), a source prior ($1.0$ for metric estimates and $0.6$ for VLM descriptions), a camera-pose factor $0.35+0.65c$, where $c$ is the lower of the heading and position confidences of its camera, each at least $0.2$, and a factor of $0.45$ for qualitative observations. A metric observation with bearing and depth takes precedence over a qualitative one for the same entity. The entity takes its position from the highest-weighted observation, and an entity observed only by cameras with an unknown position or heading is left off the layout. Its facing is the weighted vote of the Orient Anything V2 estimates over four $90^\circ$ bins. The layout uses a $10\times10$ grid, the format of the MindCube cognitive maps; the first camera is placed at the grid center facing up, and the joint camera and entity coordinates are mapped onto the grid by a monotone map that keeps interior coordinates and compresses the outlying ones. This layout is kept when the text states, with confidence at least $0.9$, that every image after the first is taken from the first camera's position with its own heading, no relations are stated, and no entity is judged to have moved with confidence at least $0.8$. Otherwise, greedy passes attempt to enforce the stated lateral and front--back orders and separate entities that share a cell. Finally, with the endpoints of satisfied stated relations fixed, a deterministic local search moves the other entities to maximize the total weight of satisfied pairwise relations minus that of violated ones: stated relations ($20$), in-front-of orders from image descriptions ($3.2$), left--right orders from detector bearings that differ by more than $15^\circ$ ($2.5$), and left--right orders from image descriptions ($1.6$), with a term of $0.5$ that keeps the queried entities apart. A cost of $0.1$ per cell of displacement from the initial placement keeps that placement in ties.

\paragraph{State passed to the Reasoner.}
The Reasoner receives the layout as a map listing each entity's position and facing and each camera's position and heading, with unknown pose components left empty. The observation level follows as text: the stated relations, the per-view observations when there are multiple images, the estimated camera motions with flags where the sources disagree, the unknown pose components, observed entities left off the map, and verdicts on whether an entity moved, for example when its positions across views disagree beyond what the camera poses explain. For questions that order distances, the distances computed from the map are appended.

\paragraph{Metric state for quantitative questions.}
\label{app:exp-metric-state}
SpatialRGPT-Bench asks for distances and sizes in metres about regions given as masks in a single image. For these questions, the given regions replace detection, and MoGe-2 back-projects the pixels of each region into 3D. The state records, in the camera frame and in metres, each region's center, its height along the estimated vertical direction, its width as the longer side of its horizontal footprint, and the center, horizontal, vertical, and surface distances between regions. The Reasoner receives these values as text, and the engine represents each region by a box at its measured center with its extent along the camera axes, with the first rendered view taken from the original camera and its field of view widened when needed to include all regions. The engine operations then use metres instead of grid cells.

\subsection{Reasoner and MuJoCo Interface}
\label{app:mujoco-interface}

Listing~\ref{lst:sandbox} gives all methods exposed to the Reasoner through the \texttt{mujoco\_sim} object, grouped as in Section~\ref{sec:method:engine}. The signatures retain the implementation's parameter defaults and return types; method bodies are omitted. Positions and distances use the units of the world state, and headings and rotation angles are in degrees. \texttt{capture\_image} returns the path of the rendered image, and its tool response lists the entities whose proxies project into the view, including occluded ones, with their distances.

\begin{nolinenumbers}
\begin{lstlisting}[
    language=Python,
    label=lst:sandbox,
    caption={Complete MuJoCo API exposed to the Reasoner through \texttt{mujoco\_sim}.},
    basicstyle=\ttfamily\footnotesize\color{py-fg},
    backgroundcolor=\color{py-bg},
    keywordstyle=\color{py-keyword}\bfseries,
    keywordstyle={[2]\color{py-builtin}},
    stringstyle=\color{py-string},
    commentstyle=\color{py-comment}\itshape,
    morekeywords={[2]Optional,Dict,List,Any,str,int,float,bool,MuJoCoSpatialExecutor},
    frame=single, framerule=0.4pt,
    rulecolor=\color{black!25},
    numbers=none,
    xleftmargin=1em, framexleftmargin=0.8em,
    breaklines=true, showstringspaces=false
]
class MuJoCoSpatialExecutor:
    """Public methods exposed through mujoco_sim."""

    # Viewpoint control
    def set_camera(self, position, facing): ...
    def rotate_left(self, angle=45): ...
    def rotate_right(self, angle=45): ...
    def turn_left(self): ...
    def turn_right(self): ...
    def turn_around(self): ...
    def move_forward(self, distance=1): ...
    def move_backward(self, distance=1): ...
    def move_left(self, distance=1): ...
    def move_right(self, distance=1): ...

    # Measurement
    def get_current_view(self) -> dict[str, list[float]]: ...
    def get_distances(self) -> dict[str, float]: ...
    def describe_camera_motion(
        self, start_view_name, target_view_name
    ) -> str: ...

    # Rendering
    def capture_image(self) -> str: ...
    def capture_image_and_metadata(self) -> str: ...  # alias of capture_image
\end{lstlisting}
\end{nolinenumbers}

\paragraph{Engine settings.}
MuJoCo instantiates labeled object proxies and executes the spatial operations requested by the Reasoner. Camera movements persist within a trajectory, and the engine is reset between questions. The Reasoner takes at most five turns. A turn without an engine call is rejected until the engine has been called once, except at the last turn, and if no answer has been given after the last turn, one final request asks for the answer without tool use.

For grid maps, the engine renders $800\times600$ RGB views with a $90^\circ$ horizontal field of view and represents objects as cylinders of radius $0.3$\,m and height $1.5$\,m. The proxy dimensions are for visualization and are not measurements of the real objects. Metric states use the boxes above.

\paragraph{Routing.}
Not every question should be answered through the engine. The world state describes positions, headings, and camera poses, and the distances and directions that follow from them; a question about appearance or occlusion lies outside this description and cannot be answered from it. Between perception and reasoning, a deterministic routing step therefore checks that $q$ is representable and that $\hat{s}$ contains what the answer needs: (i) every information type that $q$ requires, as identified by the text channel (Section~\ref{sec:method:perceiver}), must be among those the state describes; (ii) $\hat{s}$ must load in the engine; (iii) every entity in $\mathcal{O}$ must be present in $\hat{s}$; and (iv) the camera pose components that the required information types depend on must be measured. If any check fails, the engine is not used, and the VLM answers from the images and the original question. Questions for which the text channel extracts no requirement stay on the engine path, so an extraction miss cannot disable the engine. 

\subsection{Baseline Implementation}
\label{app:baselines}

In each comparison, all baselines use the same VLM as Spatial Harness and receive the same images and question; none of them receives the world state estimated by the Perceiver.

\paragraph{Base.}
The VLM answers directly from the images and question, on MindCube with the prompt in Figure~\ref{fig:prompt-raw-qa}. In the GT-map setting, the GT map is added to the prompt as text (Figure~\ref{fig:prompt-cgmap-in}). The SAT result with Qwen2.5-VL-72B in the submitted version ($60.27\%$) was affected by a scoring error; Table~\ref{tab:mainresults-crossbenchmark} reports the corrected value ($68.60\%$).

\paragraph{MindJourney.}
We use the official implementation of \textit{MindJourney}~\citep{yang2025mindjourney}, with Stable Virtual Camera~\citep{zhou2025stable} as the video world model. The VLM proposes camera trajectories, the world model generates views along them, and a spatial beam search keeps the views that the VLM judges helpful; the VLM then answers from these views and the original images. We keep the search settings of the original paper: a search depth of $3$, a beam width of $2$, and a threshold of $8$ for both the exploration and helpfulness scores. Each action moves the camera forward by $0.75$\,m or rotates it by $27^\circ$, and a trajectory moves at most $1.5$\,m forward and turns at most $60^\circ$. Stable Virtual Camera generates $8$ views per trajectory with $20$ diffusion steps and a guidance scale of $4$, and each search layer keeps at most $18$ helpful views.

\paragraph{pySpatial.}
\textit{pySpatial}~\citep{luo2026pySpatial} lets the VLM write Python programs that call 3D reconstruction, camera-pose, and novel-view tools, and the final answering stage receives the original images together with the program output. The official code was updated after our submission; all \textit{pySpatial} results in this paper use the updated release, which scores higher than the earlier release used in the submitted version (for example, $63.62\%$ instead of $59.10\%$ on MindCube with Qwen3.5-27B).

\paragraph{GCA.}
\textit{GCA}~\citep{chen2025geometrically} formalizes the question as geometric constraints and solves them with geometric tools on a scene reconstructed from the images. We run its official implementation with the same VLM on MindCube (Appendix~\ref{app:gca}).

\subsection{Prompt Templates}
\label{sec:prompts}

We show five prompt templates: the \textit{Base} prompt without and with the GT map, the Reasoner prompt for MuJoCo tool calls, the cognitive-map generation prompt used in the VLM-generated-map ablation, and the LLM-as-judge prompt for scoring. In every
figure, blue spans mark runtime substitutions and gray italics mark
conditional blocks or explanatory notes.

\paragraph{Base prompt.}

The template in Fig.~\ref{fig:prompt-raw-qa} contains a brief task statement, an answer-format instruction, and the question, paired with the input images; it is the \textit{Base} prompt on MindCube in the end-to-end setting. On SAT, MMSI-Bench, and SpatialRGPT-Bench, \textit{Base} receives the question with the benchmark's own answer-format instruction. For multiple-choice questions, when a \textit{Base} reply selects no option, one further request asks for a single option letter.

\begin{figure}[htbp!]
\centering
\begin{tcolorbox}[title = {Base Prompt (end-to-end setting)}]
\footnotesize
\texttt{[Task]}\\
\texttt{Your task is to analyze the spatial arrangement of objects in the scene by examining the provided images, which show the scene from different viewpoints.}\\

\texttt{[Answer Instruction]}\\
\texttt{You only need to provide *ONE* correct answer selecting from the options listed below. For example, if you think the correct answer is 'A. Above' from 'A. Above B. Under C. Front D. Behind', your response should **only** be '<answer>A. Above</answer>'.}\\

\texttt{[Question]}\\
\texttt{\textcolor{blue}{\{question\}}}
\end{tcolorbox}
\vspace{-7pt}
\caption{\textit{Base} prompt template on MindCube in the end-to-end setting.}
\label{fig:prompt-raw-qa}
\end{figure}

\paragraph{Base prompt with the GT map.}

The template in Fig.~\ref{fig:prompt-cgmap-in} extends the \textit{Base} prompt with a \texttt{[Cognitive Map Format]} block that declares the $10{\times}10$ bird's-view grid convention and inserts the GT map; it is the \textit{Base} prompt in the GT-map setting.

\begin{figure}[htbp!]
\centering
\begin{tcolorbox}[title = {Base Prompt (GT-map setting)}]
\footnotesize
\texttt{[Task]}\\
\texttt{Your task is to analyze the spatial arrangement of objects in the scene by examining the provided images, which show the scene from different viewpoints. Also, we provide you a cognitive map that shows the general layout for the scene. Please use the cognitive map to reason and answer the question.}\\

\texttt{[Answer Instruction]}\\
\texttt{You only need to provide *ONE* correct answer selecting from the options listed below. For example, if you think the correct answer is 'A. Above' from 'A. Above B. Under C. Front D. Behind', your response should **only** be '<answer>A. Above</answer>'.}\\

\texttt{[Cognitive Map Format]}\\
\texttt{We provide you a 2D grid map of the scene that is related to the question you should answer. Below is the description of the map:}\\
\texttt{- The map uses a 10x10 grid where [0,0] is at the top-left corner and [9,9] is at the bottom-right corner}\\
\texttt{- The map is shown in the bird's view}\\
\texttt{- Directions are defined as:}\\
\texttt{~~* up = towards the top of the grid (decreasing y-value)}\\
\texttt{~~* right = towards the right of the grid (increasing x-value)}\\
\texttt{~~* down = towards the bottom of the grid (increasing y-value)}\\
\texttt{~~* left = towards the left of the grid (decreasing x-value)}\\
\texttt{~~* inner = straight into the 2D map (perpendicular to the grid, pointing away from you)}\\
\texttt{~~* outer = straight out of the 2D map (perpendicular to the grid, pointing towards you)}\\
\texttt{- "objects" lists all important items in the scene with their positions}\\
\texttt{- "facing" indicates which direction an object is oriented towards (when applicable)}\\
\texttt{- "views" represents the different camera viewpoints in the scene}\\

\texttt{Below is the cognitive map of the scene related to the question. Please use it to reason and answer the question.}\\
\texttt{\textcolor{blue}{\{cognitive map (JSON object with "objects" and "views" arrays)\}}}\\

\texttt{[Question]}\\
\texttt{\textcolor{blue}{\{question\}}}
\end{tcolorbox}
\vspace{-7pt}
\caption{\textit{Base} prompt template in the GT-map setting.}
\label{fig:prompt-cgmap-in}
\end{figure}

\paragraph{Reasoner prompt.}

The Reasoner prompt drives the multi-turn tool-using loop over the instantiated state and consists of a top-level template with three injected blocks: the API specification, the tool-use guide, and the scene note.

The top-level template, shown in Fig.~\ref{fig:prompt-ours-template},
declares the Reasoner's role, exposes the question, references the three
injected blocks, and closes with output-format instructions. Two variants are emitted: \texttt{NO\_HIS} at step~$0$ and \texttt{WITH\_HIS} at step~$\geq 1$, differing only by the gray-bracketed history block.

\begin{figure}[htbp!]
\centering
\begin{tcolorbox}[title = {Reasoner Prompt Template}]
\footnotesize
\texttt{You are a spatial reasoning agent. Solve the problem step-by-step using the MuJoCo simulation tool.}\\

\texttt{\#\# Your Question}\\
\texttt{\textcolor{blue}{\{question\}}}\\
\textcolor{gray}{\textit{[The question is preceded by one image placeholder per input image. In the GT-map setting it contains the GT map in the format of Fig.~\ref{fig:prompt-cgmap-in}. For end-to-end grid-map inputs it contains the Perceiver's map in the same format, without the \texttt{inner}/\texttt{outer} directions and with two added notes on absolute camera headings and viewpoint-relative direction terms, followed by the observation-level text (Appendix~\ref{app:perceiver}). Once the engine has rendered a view, the latest view is appended. SpatialRGPT-Bench instead uses continuous measurements and metric versions of the API, guide, and scene note (omitted here).]}}\\

\texttt{\#\# API Reference}\\
\texttt{\textcolor{blue}{\{MuJoCo API Specification --- see Fig.~\ref{fig:prompt-ours-api}\}}}\\

\texttt{\#\# MuJoCo Tool Use Guide}\\
\texttt{\textcolor{blue}{\{MuJoCo Tool Use Guide --- see Fig.~\ref{fig:prompt-ours-tool-use}\}}}\\

\texttt{\#\# Scene Note}\\
\texttt{\textcolor{blue}{\{MuJoCo Scene Representation Note --- see Fig.~\ref{fig:prompt-ours-scene-note}\}}}\\

\textcolor{gray}{\textit{[The following block appears only in the \texttt{WITH\_HIS} variant (step $\geq 1$); the \texttt{NO\_HIS} variant used at step $0$ skips it entirely.]}}\\
\texttt{\#\# Interaction History}\\
\texttt{Prior to this step, you have already taken \textcolor{blue}{\{step\_count\}} of your \textcolor{blue}{\{max\_steps\}} step(s). Below is the interaction history}\\
\texttt{\textcolor{blue}{\{memory\_context\}}}\\
\textcolor{gray}{\textit{[End \texttt{WITH\_HIS}-only block.]}}\\

\texttt{\#\# Instructions}\\
\texttt{Now it's your turn to respond to the current step.}\\

\texttt{You should first conduct the reasoning process. This process MUST be enclosed within <think> </think> tags.}\\
\texttt{**You MUST use the MuJoCo tool at least once before giving your final answer.** Do NOT answer directly without interacting with the simulation first.}\\
\texttt{You have a budget of \textcolor{blue}{\{max\_steps\}} steps; each response is one step. After the last step you must give the final answer without further tool use.}\\
\texttt{After completing your reasoning, choose only one of the following actions (do not perform both):}\\
\texttt{1) <mujoco\_code>...</mujoco\_code>: You MUST interact with the MuJoCo simulation to gather spatial information (e.g. capture new views, measure distances). Emit exactly ONE <mujoco\_code>...</mujoco\_code> block with pure Python 3 based on the API Reference. The execution result will be returned in <tool\_response>; if it disagrees with your reasoning, correct yourself.}\\
\texttt{2) <answer>...</answer>: Only after you have already used the MuJoCo tool in a previous step, provide the self-contained solution inside <answer>...</answer>.}
\end{tcolorbox}
\vspace{-7pt}
\caption{Top-level prompt of the Reasoner.}
\label{fig:prompt-ours-template}
\end{figure}

The API specification, shown in Fig.~\ref{fig:prompt-ours-api}, enumerates
the \texttt{mujoco\_sim} methods available to the Reasoner, grouped into camera
navigation and scene observation, with a one-line contract each; Figure~\ref{fig:api} summarizes this interface, and Listing~\ref{lst:sandbox} also has the alias \texttt{capture\_image\_and\_metadata}.

\begin{figure}[htbp!]
\centering
\begin{tcolorbox}[title = {MuJoCo API Specification (block injected into Fig.~\ref{fig:prompt-ours-template})}]
\footnotesize
\texttt{Available API (a `mujoco\_sim` object and the `math` module are pre-imported; you may also use Python builtins like print/len)}\\

\texttt{Camera Navigation:}\\
\texttt{mujoco\_sim.set\_camera(position, facing)~~~~-- Teleport camera to a cognitive-map pose. position: [x, y] grid coordinates from the map or derived from the question; facing: the absolute heading from the cognitive map, either "up", "down", "left", "right", or a numeric yaw in degrees (0=right, 90=up, 180=left, 270=down; positive yaw turns left). Preserve numeric yaw without rounding. This argument is not a relative turn angle; use rotate\_left/right after setting the reference pose for relative turns. Example: mujoco\_sim.set\_camera([5, 6], "up").}\\
\texttt{mujoco\_sim.turn\_left()~~~~~~~~~~~~~~~~~~~~~-- Turn 90 degrees left from the current heading. Returns: None}\\
\texttt{mujoco\_sim.turn\_right()~~~~~~~~~~~~~~~~~~~~-- Turn 90 degrees right from the current heading. Returns: None}\\
\texttt{mujoco\_sim.turn\_around()~~~~~~~~~~~~~~~~~~~-- Turn 180 degrees (useful for "what is behind me?"). Returns: None}\\
\texttt{mujoco\_sim.rotate\_left(angle)~~~~~~~~~~~~~~-- Rotate left by an ARBITRARY angle in degrees (not limited to 90). angle: float, e.g. 44. Use this whenever the question asks about a non-right-angle turn. Returns: None}\\
\texttt{mujoco\_sim.rotate\_right(angle)~~~~~~~~~~~~~-- Rotate right by an ARBITRARY angle in degrees (not limited to 90). angle: float, e.g. 44. Use this whenever the question asks about a non-right-angle turn. Returns: None}\\
\texttt{mujoco\_sim.move\_forward(distance=1)~~~~~~~~-- Move camera forward. distance: float, grid steps (default 1). Returns: None}\\
\texttt{mujoco\_sim.move\_backward(distance=1)~~~~~~~-- Move camera backward. distance: float, grid steps. Returns: None}\\
\texttt{mujoco\_sim.move\_left(distance=1)~~~~~~~~~~~-- Strafe camera left (facing unchanged). distance: float, grid steps. Returns: None}\\
\texttt{mujoco\_sim.move\_right(distance=1)~~~~~~~~~~-- Strafe camera right (facing unchanged). distance: float, grid steps. Returns: None}\\

\texttt{Observation:}\\
\texttt{mujoco\_sim.capture\_image()~~~~~~~~~~~~~~~~-- Take a photo. Returns: str (image file path). The tool response also includes [capture\_info].objects\_in\_view, a list of \{"name": str, "distance": float\}; use those names/distances as evidence, not the path string itself.}\\
\texttt{mujoco\_sim.get\_current\_view()~~~~~~~~~~~~~-- Get the camera's current pose. Returns: \{"location": [x, y, z], "rotation": [pitch, yaw, roll]\}. x and y are the camera's cognitive-map grid coordinates and z is its height; rotation is in degrees and yaw uses the same convention as set\_camera's facing. Read yaw as pose["rotation"][1]. There are no "position" or "facing" keys. Example: pose = mujoco\_sim.get\_current\_view(); x, y = pose["location"][:2]; yaw\_deg = pose["rotation"][1]}\\
\texttt{mujoco\_sim.get\_distances()~~~~~~~~~~~~~~~~-- Get distances from the camera to each object in grid units. Returns: dict, e.g. \{"object\_a": 3.5, "object\_b": 5.2\} (names match the cognitive map exactly)}\\
\texttt{mujoco\_sim.describe\_camera\_motion(v1, v2)~~-- Describe camera/observer motion from view v1 to v2; this does not prove that an object itself moved. v1, v2: str. Returns: str, e.g. "Forward", "Left", "Diagonally forward and right"}\\
\end{tcolorbox}
\vspace{-7pt}
\caption{MuJoCo API specification given to the Reasoner.}
\label{fig:prompt-ours-api}
\end{figure}

The tool-use guide, shown in Fig.~\ref{fig:prompt-ours-tool-use}, restricts the Reasoner to the listed API, asks it to execute the queried
move or turn before capturing, to re-orient when \texttt{capture\_image}
returns no relevant objects, and to derive the answer from the latest tool
response, followed by worked examples.

\begin{figure}[htbp!]
\centering
\begin{tcolorbox}[title = {MuJoCo Tool Use Guide (block injected into Fig.~\ref{fig:prompt-ours-template})}]
\footnotesize
\texttt{(To save context, we do not provide images in the examples)}\\
\texttt{Use only the APIs listed in the API Reference above.}\\
\texttt{For "what is to my left/right/behind after ..." questions, first execute the described movement or turn, then turn the camera toward the queried side and call capture\_image(); equal get\_distances() values alone do not identify direction.}\\
\texttt{If capture\_image() returns an empty objects\_in\_view list or no answer-option objects, change to the relevant adjacent facing and capture again before answering.}\\
\texttt{Answer from the tool response: before answering, state in <think> what the latest tool response shows (the objects\_in\_view names, distances or printed values) and derive the answer from it.}\\

\texttt{Example Question 1:}\\
\texttt{Based on these four images (image 1, 2, 3, and 4) showing the black sneaker from different viewpoints (front, left, back, and right), with each camera aligned with room walls and partially capturing the surroundings: From the viewpoint presented in image 2, what is to the right of the black sneaker?}\\

\texttt{Example Answer 1:}\\
\texttt{<think>To determine what is located to the right of the black sneaker from the perspective of Image 2, I need to adjust the camera's orientation. First, I will find Image 2 in the cognitive map's views list and set the camera using that view's position and facing. Since I need to identify what is to the right of the sneaker from this specific vantage point, I will turn the camera right to face the queried side. Finally, I will capture a new image to visually identify the surrounding objects in that direction.</think>}\\
\texttt{<mujoco\_code>}\\
\texttt{\# From the cognitive map, Image 2 has position [4, 5] and facing "right".}\\
\texttt{mujoco\_sim.set\_camera([4, 5], "right")}\\
\texttt{mujoco\_sim.turn\_right()}\\
\texttt{image\_path = mujoco\_sim.capture\_image()}\\
\texttt{</mujoco\_code>}\\
\end{tcolorbox}
\caption{Tool-use guide given to the Reasoner.}
\label{fig:prompt-ours-tool-use}
\end{figure}

\begin{figure}[p]
\ContinuedFloat
\centering
\begin{tcolorbox}[title = {MuJoCo Tool Use Guide (continued)}]
\footnotesize
\texttt{Example Question 2:}\\
\texttt{Based on these four images (image 1, 2, 3, and 4) showing the pink bottle from different viewpoints (front, left, back, and right), with each camera aligned with room walls and partially capturing the surroundings: If I am standing at the same spot and facing the same direction as shown in image 1, then I turn right and move forward, will I get closer to the pink plush toy and headboard?}\\

\texttt{Example Answer 2:}\\
\texttt{<think>To determine if turning right and moving forward from the viewpoint of Image 1 brings me closer to the pink plush toy and the headboard, I need to measure the change in distance to these objects. First, I will initialize the camera to the position and orientation of Image 1. Before moving, I will retrieve the current distances to all objects in the scene to establish a baseline. Next, I will perform the actions specified in the prompt: rotating 90 degrees to the right and then moving forward. Finally, I will call the distance measurement function again. By comparing the distance to the pink plush toy and headboard before and after the movement, I can quantitatively verify whether the proximity has increased.</think>}\\
\texttt{<mujoco\_code>}\\
\texttt{\# From the cognitive map, Image 1 has position [5, 6] and facing "up".}\\
\texttt{mujoco\_sim.set\_camera([5, 6], "up")}\\
\texttt{distance\_before = mujoco\_sim.get\_distances()}\\
\texttt{mujoco\_sim.turn\_right()}\\
\texttt{mujoco\_sim.move\_forward()}\\
\texttt{distance\_after = mujoco\_sim.get\_distances()}\\
\texttt{print("before", distance\_before)}\\
\texttt{print("after", distance\_after)}\\
\texttt{</mujoco\_code>}\\

\textcolor{gray}{\textit{[Three further worked examples follow in the guide and are omitted here: moving to a marked location and comparing distances, identifying the camera motion between two views with \texttt{describe\_camera\_motion}, and a two-step example in which the answer is read from the returned \texttt{objects\_in\_view}.]}}
\end{tcolorbox}
\vspace{-7pt}
\caption{Tool-use guide given to the Reasoner (continued).}
\end{figure}

The scene note, shown in Fig.~\ref{fig:prompt-ours-scene-note}, calibrates
the Reasoner to MuJoCo's stylized renderings and to the
\texttt{set\_camera(position, facing)} initialization from the map.

\begin{figure}[htbp!]
\centering
\begin{tcolorbox}[title = {MuJoCo Scene Representation Note (block injected into Fig.~\ref{fig:prompt-ours-template})}]
\footnotesize
\texttt{Objects in the images are shown as labeled cylinders (placeholders), NOT real object appearances. Focus on spatial relationships (positions, distances, directions) between them.}\\

\texttt{The question may include a cognitive map. To initialize a camera pose, use the relevant [x, y] position and facing from the map or derive them from the question. For an existing camera view, use that view's position and facing. For a target point such as "X" or another marked location, use that point's position and the intended facing direction.}
\end{tcolorbox}
\vspace{-7pt}
\caption{Scene note given to the Reasoner: it describes MuJoCo's stylized renderings (colored cylinders with labels rather than photo-realistic objects) and the camera initialization from the map.}
\label{fig:prompt-ours-scene-note}
\end{figure}

\paragraph{Cognitive-Map Generation.}

The cognitive-map generation template, shown in
Fig.~\ref{fig:prompt-cogmap-gen}, is invoked once per question in the
VLM-generated-map ablation (Table~\ref{tab:ablation-mindcube}).

\begin{figure}[htbp!]
\centering
\begin{tcolorbox}[title = {Cognitive-Map Generation Prompt}]
\footnotesize
\texttt{[Task]}\\
\texttt{Your task is to analyze the spatial arrangement of objects in the scene by examining the provided images, which show the scene from different viewpoints. You will then create a detailed cognitive map representing the scene using a 10x10 grid coordinate system.}\\

\texttt{[Rules]}\\
\texttt{Create a cognitive map with the following structure in the bird's view:}\\
\texttt{~~~- A 10x10 grid where [0,0] is at the top-left corner and [9,9] is at the bottom-right corner}\\
\texttt{~~~- up = towards the top of the grid (decreasing y)}\\
\texttt{~~~- right = towards the right of the grid (increasing x)}\\
\texttt{~~~- down = towards the bottom of the grid (increasing y)}\\
\texttt{~~~- left = towards the left of the grid (decreasing x)}\\
\texttt{~~~- inner = straight into the 2D map (perpendicular to the grid, pointing away from you)}\\
\texttt{~~~- outer = straight out of the 2D map (perpendicular to the grid, pointing towards you)}\\
\texttt{~~~- Include positions of all objects from the specified categories}\\
\texttt{~~~- Estimate the center location (coordinates [x, y]) of each instance within provided categories}\\
\texttt{~~~- If a category contains multiple instances, include all of them}\\
\texttt{~~~- Each object's estimated location should accurately reflect its real position in the scene, preserving the relative spatial relationships among all objects}\\
\texttt{~~~- Combine and merge information from the images since they are pointing to the same scene, calibrating the object locations accordingly}\\
\texttt{~~~- Include camera positions and directions for each view}\\
\texttt{3. Carefully integrate information from all views to create a single coherent spatial representation.}\\

\texttt{[Output Format]}\\
\texttt{Return your cognitive map and nothing else, in the following exact format:}\\

\texttt{<cogmap>}\\
\texttt{\{}\\
\texttt{~~"objects": [}\\
\texttt{~~~~\{"name": "object\_name", "position": [x, y], "facing": "direction"\},}\\
\texttt{~~~~\{"name": "object\_without\_orientation", "position": [x, y]\}}\\
\texttt{~~],}\\
\texttt{~~"views": [}\\
\texttt{~~~~\{"name": "Image 1", "position": [x, y], "facing": "direction"\},}\\
\texttt{~~~~\{"name": "Image 2", "position": [x, y], "facing": "direction"\}}\\
\texttt{~~]}\\
\texttt{\}}\\
\texttt{</cogmap>}\\

\texttt{Do NOT include reasoning, commentary, or an answer to any question. Output ONLY the <cogmap>...</cogmap> block.}\\

\textcolor{gray}{\textit{[The text block above is followed in the same user message by the scene's $N$ multi-view images, sent as base64-encoded \texttt{image\_url} parts in the canonical \texttt{Image $1, 2, \dots, N$} order recorded in the dataset's \texttt{images} field, and then by the question with its options.]}}\\

\texttt{[Question]}\\
\texttt{\textcolor{blue}{\{question\}}}
\end{tcolorbox}
\vspace{-7pt}
\caption{Prompt used to elicit a cognitive map from a VLM.}
\label{fig:prompt-cogmap-gen}
\end{figure}

\paragraph{Prompt for LLM-as-Judge.}

The LLM-as-judge template, shown in Fig.~\ref{fig:prompt-judge}, scores
the final answer against the reference answer on the multiple-choice
benchmarks (Appendix~\ref{app:exp-evaluation}).

\begin{figure}[htbp!]
\centering
\begin{tcolorbox}[title = {LLM-as-Judge Prompt}]
\footnotesize
\texttt{You are a strict answer-correctness judge. Your job is to determine whether the model's FINAL ANSWER TEXT explicitly matches the ground-truth answer for a given question.}\\

\texttt{\#\# Question}\\
\texttt{\textcolor{blue}{\{question\}}}\\

\texttt{\#\# Ground-Truth Answer}\\
\texttt{\textcolor{blue}{\{ground\_truth\}}}\\

\texttt{\#\# Model's Final Answer}\\
\texttt{\textcolor{blue}{\{final\_answer\_text\}}}\\

\texttt{\#\# Instructions}\\
\texttt{1. Judge ONLY the model's final answer text shown above. Do NOT solve the task from scratch, do NOT use the cognitive map or question details to infer a missing answer, and do NOT give credit for reasoning that is not actually stated in the final answer text.}\\
\texttt{2. The ground-truth answer is given as a letter (e.g., A, B, C, D) or a short text representing the correct option.}\\
\texttt{3. The model's final answer may contain the option letter, the option text, or both. Mark "correct" only if the final answer text itself clearly selects the same option as the ground truth.}\\
\texttt{4. If the final answer text is unresolved, templated, or not an actual final answer, mark "incorrect".}\\
\texttt{5. If the final answer text is empty, ambiguous, or does not clearly commit to one final option, mark "incorrect".}\\
\texttt{6. Focus on what the model explicitly answered, not what you think the correct answer should be.}\\

\texttt{\#\# Output format (STRICT)}\\
\texttt{Respond with ONLY a single JSON object on one line, with no extra text:}\\
\texttt{\{"verdict":"correct"\}}\\
\texttt{or}\\
\texttt{\{"verdict":"incorrect"\}}
\end{tcolorbox}
\vspace{-7pt}
\caption{LLM-as-judge prompt for scoring free-form answers.}
\label{fig:prompt-judge}
\end{figure}

\subsection{Evaluation Setup}
\label{app:exp-data}
\label{app:experimental-details}
\label{app:exp-implementation}

\paragraph{Datasets.}
MindCube-Tiny groups its $1{,}050$ questions by camera motion: \textit{Rotation} ($200$ questions), where the camera rotates in place; \textit{Among} ($600$), where the views face a central object from different sides; and \textit{Around} ($250$), where the camera moves around a group of objects. Its GT maps are used in the GT-map comparison, the map-quality measurements, and the map-corruption analysis. No method receives them in end-to-end inference.

SAT asks about the spatial effects of ego movement, object movement, action consequences, goal aiming, and perspective changes. MindJourney generates videos for every question, so evaluating all $4{,}001$ questions of the SAT validation split is expensive. We therefore sample $100$ questions from each of the five categories with seeds $42$ to $46$, one per category, giving $500$ questions (Table~\ref{tab:mainresults-sat-detail}).

MMSI-Bench groups its $1{,}000$ questions into positional relationship ($522$; its six positional question types), attribute ($130$; measurement and appearance), motion ($150$; camera and object motion), and multi-step reasoning ($198$) (Table~\ref{tab:mainresults-mmsi}). The quantitative split of SpatialRGPT-Bench refers to objects by image regions. Its $749$ questions ask for direction ($107$), distance ($148$), height ($133$), horizontal distance ($122$), vertical distance ($106$), and width ($133$) (Table~\ref{tab:mainresults-spatialrgpt}).

Within each comparison, all methods use the same questions. Benchmark category labels are used only to organize the results and are not given to Spatial Harness, except on SpatialRGPT-Bench, where the category selects the requested answer format (clock hour or metres).

\begin{figure*}
\begin{center}
\includegraphics[width=\textwidth]{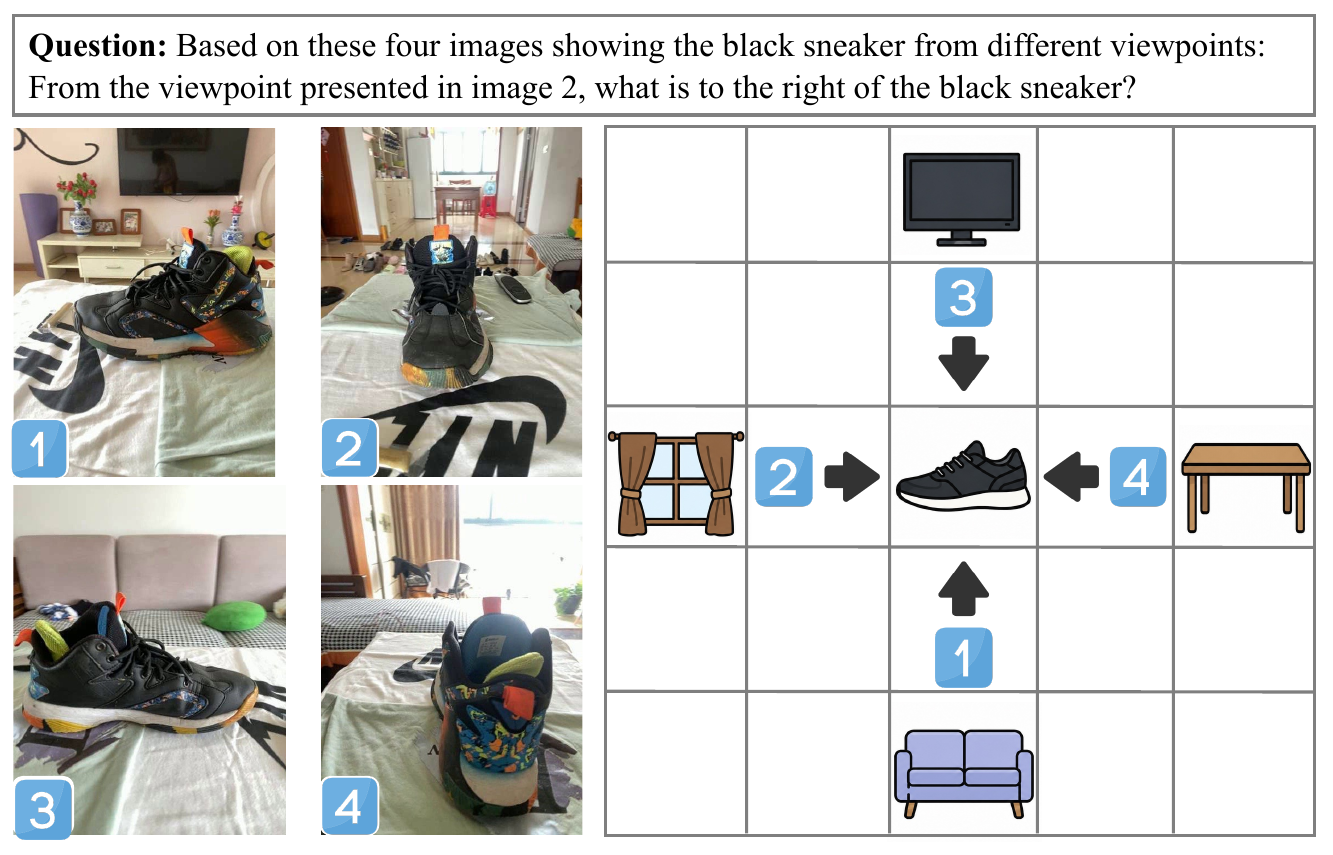}
\end{center}
\caption{
An example from the MindCube dataset. The top shows the question, the left presents four viewpoint images, and the right visualizes the corresponding cognitive map (see Figure~\ref{fig:cogmap-json-example}).
}
\label{fig:cogmap-example}
\end{figure*}

\paragraph{Ground-truth states.}
For Table~\ref{tab:mainresults-mindcube-map}, the annotated map provides objects, spatial configurations, and capture-camera poses. \textit{Base} (w map) reasons directly from the annotation, Spatial Harness (w map) instantiates it in the engine, and \textit{Base} (w/o map) receives only the images and question. The same annotated-state setting supplies the GT reference in Table~\ref{tab:ablation-mindcube} and the fixed states in Table~\ref{tab:ablation-backend}. GT-map performance is a reference condition, not a strict upper bound on answer accuracy. Figures~\ref{fig:cogmap-example} and~\ref{fig:cogmap-json-example} show an example GT map.

\begin{figure}[htbp!]
\centering
\begin{tcolorbox}[title = {Example Cognitive Map}]
\footnotesize
\texttt{\{}\\
\texttt{~~"objects": [}\\
\texttt{~~~~\{"name": "black sneaker",~~~~~~~~~~~~~~~"position": [5, 5], "facing": "left"\},}\\
\texttt{~~~~\{"name": "light purple sofa",~~~~~~~~~~~"position": [5, 8], "facing": "down"\},}\\
\texttt{~~~~\{"name": "brown curtains and windows", "position": [2, 5]\},}\\
\texttt{~~~~\{"name": "TV",~~~~~~~~~~~~~~~~~~~~~~~~~~"position": [5, 2]\},}\\
\texttt{~~~~\{"name": "wooden dining table",~~~~~~~~~"position": [8, 5]\}}\\
\texttt{~~],}\\
\texttt{~~"views": [}\\
\texttt{~~~~\{"name": "Image 1", "position": [5, 6], "facing": "up"\},}\\
\texttt{~~~~\{"name": "Image 2", "position": [4, 5], "facing": "right"\},}\\
\texttt{~~~~\{"name": "Image 3", "position": [5, 4], "facing": "down"\},}\\
\texttt{~~~~\{"name": "Image 4", "position": [6, 5], "facing": "left"\}}\\
\texttt{~~]}\\
\texttt{\}}
\end{tcolorbox}
\vspace{-7pt}
\caption{Example cognitive map. Five objects on a $10{\times}10$ bird's-view grid (\texttt{[0,0]} at the top-left, \texttt{[9,9]} at the bottom-right; $+x$ right, $+y$ down), and four camera views around the central \emph{black sneaker}; \texttt{facing} is one of \{\texttt{up}, \texttt{down}, \texttt{left}, \texttt{right}\}.}
\label{fig:cogmap-json-example}
\end{figure}

\paragraph{Predicted states.}
In end-to-end experiments, the Perceiver receives the images and question, including any accompanying scene description. Reference maps, answers, and reasoning traces are excluded from inference. The Reasoner answers by querying the instantiated state. Region references supplied with a benchmark question identify the queried objects; reference metric values are not substituted for predicted measurements.

MMSI-Bench uses the same Perceiver, tools, and engine API as MindCube and SAT, without benchmark-specific tuning; its motion questions concern relations between the given views and do not involve contact-dynamics rollouts. For SpatialRGPT-Bench, the state uses continuous metric coordinates to support quantitative questions (Appendix~\ref{app:exp-metric-state}).

\paragraph{Models and decoding.}
Qwen3.5-27B and Qwen2.5-VL-7B/32B/72B-Instruct are served locally with vLLM in FP8 precision (FP8 checkpoints, except Qwen2.5-VL-32B-Instruct, which is quantized to FP8 at load time); GPT-5.4-mini is queried through the OpenAI API, and Kimi K2.5 and GLM-5V-Turbo through OpenRouter. All calls use an OpenAI-compatible interface with temperature $0$, no sampling seed, and input images of at most $409{,}600$ pixels. Thinking is disabled through the chat template for Qwen3.5-27B; requests to Kimi K2.5 and GLM-5V-Turbo set OpenRouter's \texttt{reasoning.enabled} to false, and those to Kimi K2.5 also set the reasoning effort to none; GPT-5.4-mini uses its default settings. The output limit is $8{,}192$ tokens per call for \textit{Base}, the Reasoner, and the VLM-generated map, $4{,}096$ per round for multi-turn self-review, and $1{,}024$ for each Perceiver query.

\paragraph{Compute resources.}\label{app:compute}
Qwen2.5-VL-72B-Instruct is served on two NVIDIA A100 80GB GPUs and the other Qwen models on one; the closed-weight models are accessed through APIs.

\label{app:exp-evaluation}
\paragraph{Answer scoring.}
All Spatial Harness and \textit{Base} results are scored by the same LLM judge, Qwen3.5-27B in FP8 precision at temperature $0$, which returns a JSON verdict or the extracted answer values, whichever model is evaluated. For multiple-choice questions (MindCube, SAT, and MMSI-Bench), it decides whether the final answer explicitly selects the reference option with the prompt in Figure~\ref{fig:prompt-judge}. An empty answer is scored as incorrect without a judge call, and an ambiguous answer or one that selects several options is incorrect. The evaluator operates after inference and provides no feedback to the Reasoner. If the Reasoner's final request still yields no option, the VLM answers from the images and question, as in Algorithm~\ref{alg:visual-harness}. Accuracy is the fraction of questions scored as correct.

\paragraph{Quantitative questions.}
On SpatialRGPT-Bench, the judge extracts a value with its unit, or a clock direction, from the final answer. A distance or size is correct when it lies within $[0.75,1.25]$ times the reference value after conversion to metres, with a value without unit read as metres; a direction is correct when it is within one hour of the reference on a twelve-hour clock. The success rate is the fraction of correct answers and is distinct from multiple-choice answer accuracy.

\section{Additional Results}
\label{app:additional-results}

This section gives per-category results, the \textit{GCA} comparison, and details of the analyses in Section~\ref{sec:experiments}.

\subsection{Category-Level Benchmark Results}
\label{app:category-results}

Tables~\ref{tab:mainresults-sat-detail}--\ref{tab:mainresults-spatialrgpt} give the category scores corresponding to the end-to-end comparison.

\begin{table}[!t]
\centering
\caption{End-to-end accuracy (\%) on the sampled synthetic split of SAT. EgoM: ego movement; ObjM: object movement; Act: action consequence; Goal: goal aim; Pers: perspective.}
\label{tab:mainresults-sat-detail}
\fontsize{8}{9.5}\selectfont
\setlength{\tabcolsep}{2.5pt}
\begin{tabular*}{\textwidth}{@{\extracolsep{\fill}}llcccccc@{}}
\toprule
\textbf{Model} & \textbf{Method} & \textbf{EgoM} & \textbf{ObjM} & \textbf{Act} & \textbf{Goal} & \textbf{Pers} & \textbf{Overall} \\
\midrule
\multirow{4}{*}{Qwen2.5-VL-72B-Instruct} & Base & 84.00 & 70.00 & 56.00 & 71.00 & 62.00 & 68.60 \\
 & MindJourney & 86.00 & 61.00 & 57.00 & 90.00 & 49.00 & 68.60 \\
 & pySpatial & 42.00 & 41.00 & 56.00 & 73.00 & 65.00 & 55.40 \\
 & Spatial Harness & 76.00 & 70.00 & 65.00 & 82.00 & 65.00 & 71.60 \\
\midrule
\multirow{4}{*}{Qwen3.5-27B} & Base & 95.00 & 81.00 & 71.00 & 91.00 & 71.00 & 81.80 \\
 & MindJourney & 99.00 & 75.00 & 74.00 & 89.00 & 55.00 & 78.40 \\
 & pySpatial & 61.00 & 75.00 & 50.00 & 82.00 & 72.00 & 68.00 \\
 & Spatial Harness & 93.00 & 79.00 & 69.00 & 91.00 & 84.00 & 83.20 \\
\bottomrule
\end{tabular*}
\end{table}

\begin{table}[!t]
\centering
\caption{End-to-end accuracy (\%) on MMSI-Bench. Pos.: positional relationship; Attr.: attribute; Multi-step: multi-step reasoning. Overall aggregates all questions.}
\label{tab:mainresults-mmsi}
\fontsize{8}{10}\selectfont
\setlength{\tabcolsep}{3pt}
\begin{tabular*}{\textwidth}{@{\extracolsep{\fill}}llccccc@{}}
\toprule
\textbf{Model} & \textbf{Method} & \textbf{Pos.} & \textbf{Attr.} & \textbf{Motion} & \textbf{Multi-step} & \textbf{Overall} \\
\midrule
\multirow{4}{*}{Qwen2.5-VL-72B-Instruct} & Base & 30.84 & 34.62 & 24.00 & 27.27 & 29.60 \\
 & MindJourney & 31.42 & 35.38 & 28.00 & 27.27 & 30.60 \\
 & pySpatial & 33.33 & 36.15 & 34.00 & 30.30 & 33.20 \\
 & Spatial Harness & 29.89 & 34.62 & 23.33 & 27.27 & 29.00 \\
\midrule
\multirow{4}{*}{Qwen3.5-27B} & Base & 35.44 & 34.62 & 32.67 & 27.78 & 33.40 \\
 & MindJourney & 32.76 & 38.46 & 26.00 & 35.35 & 33.00 \\
 & pySpatial & 37.36 & 36.92 & 38.67 & 33.33 & 36.70 \\
 & Spatial Harness & 40.23 & 41.54 & 38.00 & 37.37 & 39.50 \\
\bottomrule
\end{tabular*}
\end{table}

\begin{table}[!t]
\centering
\caption{Quantitative success rate (\%) on SpatialRGPT-Bench. H-dist.\ and V-dist.\ denote horizontal and vertical distance. Scores use the benchmark metric and direction tolerances.}
\label{tab:mainresults-spatialrgpt}
\fontsize{8}{10}\selectfont
\setlength{\tabcolsep}{2pt}
\begin{tabular*}{\textwidth}{@{\extracolsep{\fill}}llccccccc@{}}
\toprule
\textbf{Model} & \textbf{Method} & \textbf{Dir.} & \textbf{Dist.} & \textbf{Height} & \textbf{H-dist.} & \textbf{V-dist.} & \textbf{Width} & \textbf{Overall} \\
\midrule
\multirow{4}{*}{Qwen2.5-VL-72B-Instruct} & Base & 60.75 & 25.00 & 58.65 & 33.61 & 26.42 & 42.11 & 40.72 \\
 & MindJourney & 65.42 & 25.00 & 55.64 & 31.15 & 19.81 & 39.85 & 39.12 \\
 & pySpatial & 54.21 & 14.86 & 42.86 & 22.95 & 22.64 & 36.09 & 31.64 \\
 & Spatial Harness & 98.13 & 89.86 & 75.19 & 88.52 & 86.79 & 69.17 & 84.11 \\
\midrule
\multirow{4}{*}{Qwen3.5-27B} & Base & 60.75 & 31.76 & 73.68 & 35.25 & 36.79 & 51.88 & 48.20 \\
 & MindJourney & 63.55 & 28.38 & 66.17 & 36.89 & 28.30 & 51.88 & 45.66 \\
 & pySpatial & 62.62 & 27.03 & 58.65 & 22.13 & 23.58 & 45.86 & 39.79 \\
 & Spatial Harness & 98.13 & 89.86 & 75.19 & 87.70 & 86.79 & 65.41 & 83.31 \\
\bottomrule
\end{tabular*}
\end{table}

\subsection{Comparison with GCA}
\label{app:gca}

\textit{GCA}~\citep{chen2025geometrically} formalizes the question as geometric constraints and solves them with geometric tools on a scene reconstructed from the images. We run it on MindCube under the same protocol as the end-to-end baselines in Table~\ref{tab:mainresults-mindcube}. Table~\ref{tab:gca} shows that Spatial Harness achieves the highest overall accuracy with both models. Like \textit{pySpatial}, \textit{GCA} reasons over a scene reconstructed from pixels, which cannot take in facts stated only in the question, such as the camera motion between views.

\begin{table}[!t]
\centering
\caption{End-to-end accuracy (\%) on MindCube compared with \textit{GCA}. Bold: best overall accuracy for each model; underline: second best.}
\label{tab:gca}
\fontsize{8}{10.5}\selectfont
\setlength{\tabcolsep}{4pt}
\begin{tabular*}{\textwidth}{@{\extracolsep{\fill}}llcccc@{}}
\toprule
\textbf{Model} & \textbf{Method} & \textbf{Rotation} & \textbf{Among} & \textbf{Around} & \textbf{Overall} \\
\midrule
\multirow{3}{*}{Qwen2.5-VL-72B-Instruct} & Base & 40.83 & 41.06 & 42.93 & \underline{41.46} \\
 & GCA & 39.00 & 32.67 & 31.20 & 33.52 \\
 & Spatial Harness & 49.50 & 58.00 & 53.60 & \textbf{55.33} \\
\midrule
\multirow{3}{*}{Qwen3.5-27B} & Base & 77.17 & 41.72 & 59.47 & 52.70 \\
 & GCA & 81.50 & 51.33 & 73.20 & \underline{62.29} \\
 & Spatial Harness & 81.50 & 68.00 & 70.40 & \textbf{71.14} \\
\bottomrule
\end{tabular*}
\end{table}

\subsection{World State Construction Ablations}
\label{app:exp-ablations}
\label{app:perceiver-internal-results}

\paragraph{State source and channel ablations.}
Table~\ref{tab:ablation-mindcube} holds the downstream Reasoner fixed and changes the state source. VLM-generated maps are produced directly from the images and question without the two-channel Perceiver (Figure~\ref{fig:prompt-cogmap-gen}); when a generated map fails the format check ($237$ of $1{,}050$ questions), the VLM answers directly from the images and question. Without the text channel, the VLM lists the visible objects from the images without the question, and the relations and camera poses stated in the text are dropped. Without the visual channel, it estimates the visual fields itself while retaining the text channel's structured output. Full Perceiver (Spatial Harness) uses both channels; the GT map supplies the annotated-state reference. These conditions distinguish direct map generation from replacing one channel within the structured pipeline.

\paragraph{Text-module interventions.}
Figure~\ref{fig:channel-ablation}(a) varies the structured entity-and-relation module and the text-derived camera/view constraints separately, keeping the remaining state-construction stages and Reasoner fixed. The former extracts question-relevant entities and relations; the latter recovers stated camera motion and cross-view constraints. They are separate interventions rather than a joint removal of the entire text channel: the first replaces the extracted entities with the objects the VLM lists from the images and drops the stated relations while keeping the camera constraints, and the second drops the camera constraints while keeping the entities and relations.

\paragraph{Visual-estimator substitutions.}
Each condition in Figure~\ref{fig:channel-ablation}(b) replaces one specialist with the VLM performing the same task in the same units. The detector replacement supplies regions to the unchanged depth and orientation estimators. Replacing depth, orientation, or relative pose preserves the other estimators. This substitution is necessary because deleting a parent output would also disable its dependent stages and measure the loss of an entire branch.



\subsection{Grounded Feedback Conditions}
\label{app:feedback-conditions}

Table~\ref{tab:ablation-backend} uses Qwen3.5-27B and a fixed GT map; the conditions differ in the feedback available to the Reasoner and the corresponding interaction protocol. \textit{Base} answers from the map without engine feedback. Multi-turn self-review makes a second pass over the initial answer without new scene information. Numerical feedback only exposes geometric computations without rendered images. Rendered feedback only exposes rendered views after viewpoint changes; distance queries and camera-motion descriptions are removed, and only the camera pose remains readable. Spatial Harness provides both feedback types. These comparisons isolate reasoning over a supplied state from errors in state estimation.

\subsection{World State Quality Metrics}
\label{app:exp-state-metrics}

Figure~\ref{fig:state-quality} evaluates Perceiver maps against the annotations for all $1{,}050$ MindCube questions. The four reported metrics lie in $[0,1]$:
\begin{itemize}[leftmargin=*,itemsep=2pt,topsep=3pt,parsep=0pt]
    \item \textit{Validity}: the fraction of predicted maps passing the required format checks.
    \item \textit{Coverage}: the fraction of annotated objects and views recovered under name matching.
    \item \textit{Directional Similarity}: the match rate of pairwise directional relations between recovered entities, over the pairs with an annotated relation.
    \item \textit{Facing Similarity}: facing-label accuracy for recovered entities with annotated facing.
\end{itemize}
Figure~\ref{fig:state-quality} aligns names before scoring: generated \texttt{Image}~$N$ and \texttt{View}~$N$ names are matched to the annotated views, and a generated object name is matched one-to-one to an annotated object when an LLM name judge (Qwen3.5-27B) accepts the two names as the same object. As in the official MindCube scorer, the predicted map is compared under the rotation that best aligns it with the annotation. Our implementation reproduces the official scorer (commit \texttt{6d8fd9a}) on all $1{,}050$ questions. Scores are averaged over all $1{,}050$ questions, and missing or malformed maps remain in the denominator as empty maps. These metrics measure agreement with the annotated map, not continuous metric reconstruction error. The resulting map can pass format checks while omitting entities or assigning incorrect spatial relations.

\subsection{Sensitivity to State Errors}
\label{app:state-noise-results}

Table~\ref{tab:map-noise} corrupts one field type of the GT map at a time while holding Qwen3.5-27B and the Reasoner fixed. Starting from GT maps separates the injected error from errors already present in predicted maps. The uncorrupted GT-map condition supplies the reference accuracy. Each condition is run with three corruption seeds ($42$, $43$, $44$), and the table reports the mean accuracy.

For object position, one object selected by the seed moves by one, two, or three cells along one of the eight grid directions, to a free cell inside the grid; the three levels move the same object in the same direction. For camera pose, only the yaw of one view selected by the seed changes, with perturbation magnitudes in $(0,30]$, $(30,90]$, or $(90,180]$ degrees. For object facing, one or two annotated facings rotate by $90^\circ$ to the left or right. Facing has only two severity levels because each MindCube item annotates at most two facings. Facing corruption applies only to items with enough annotated facings: $364$ questions have at least one and $129$ have two. On these questions, the uncorrupted accuracies are $97.53$ and $98.45$. All other fields remain unchanged. These interventions test separate error dimensions, not simultaneous corruptions or a fitted model of natural Perceiver errors.
\begin{table}[!t]
\centering
\caption{MindCube accuracy (\%) under single-dimension corruption of the GT map, with Qwen3.5-27B. The uncorrupted reference accuracy is $98.25$. Each row changes only the specified field. Object facing is evaluated on the $364$ and $129$ questions with at least one and two annotated facings, whose uncorrupted accuracies are $97.53$ and $98.45$.}
\label{tab:map-noise}
\label{tab:ablation-state-corruption}
\fontsize{8}{10}\selectfont
\setlength{\tabcolsep}{3pt}
\begin{tabular*}{\textwidth}{@{\extracolsep{\fill}}lccc@{}}
\toprule
\textbf{Corruption} & \textbf{Low} & \textbf{Mid} & \textbf{High} \\
\midrule
One object position & \makecell{1 cell\\92.13} & \makecell{2 cells\\89.40} & \makecell{3 cells\\87.78} \\
\midrule
One camera yaw & \makecell{$(0,30^\circ]$\\98.10} & \makecell{$(30^\circ,90^\circ]$\\84.89} & \makecell{$(90^\circ,180^\circ]$\\76.67} \\
\midrule
Object facing ($90^\circ$ rotation) & \makecell{1 object\\96.06} & N/A & \makecell{2 objects\\94.32} \\
\bottomrule
\end{tabular*}
\end{table}

\begin{table}[!t]
\centering
\caption{Run-to-run variation on MindCube in the GT-map setting. Entries are mean $\pm$ sample standard deviation of accuracy (\%) over three runs.}
\label{tab:mainresults-mindcube-map-seed}
\fontsize{8}{10}\selectfont
\setlength{\tabcolsep}{2pt}
\begin{tabular*}{\textwidth}{@{\extracolsep{\fill}}llcccc@{}}
\toprule
\textbf{Model} & \textbf{Method} & \textbf{Rotation} & \textbf{Among} & \textbf{Around} & \textbf{Overall} \\
\midrule
\multirow{3}{*}{Qwen2.5-VL-7B-Instruct} & Base (w/o map) & 35.83 $\pm$ 0.76 & 33.39 $\pm$ 0.10 & 40.67 $\pm$ 0.83 & 35.59 $\pm$ 0.29 \\
 & Base (w map) & 35.50 $\pm$ 0.50 & 27.56 $\pm$ 0.35 & 32.93 $\pm$ 0.92 & 30.35 $\pm$ 0.29 \\
 & Spatial Harness (w map) & 34.33 $\pm$ 4.31 & 58.17 $\pm$ 0.44 & 70.27 $\pm$ 2.20 & 56.51 $\pm$ 1.10 \\
\midrule
\multirow{3}{*}{Qwen2.5-VL-32B-Instruct} & Base (w/o map) & 38.83 $\pm$ 0.29 & 38.17 $\pm$ 0.29 & 46.00 $\pm$ 0.40 & 40.16 $\pm$ 0.22 \\
 & Base (w map) & 40.83 $\pm$ 2.47 & 40.11 $\pm$ 0.51 & 65.73 $\pm$ 0.61 & 46.35 $\pm$ 0.38 \\
 & Spatial Harness (w map) & 69.00 $\pm$ 1.00 & 88.39 $\pm$ 0.54 & 84.53 $\pm$ 1.15 & 83.78 $\pm$ 0.71 \\
\midrule
\multirow{3}{*}{Qwen2.5-VL-72B-Instruct} & Base (w/o map) & 40.83 $\pm$ 0.29 & 41.06 $\pm$ 0.35 & 42.93 $\pm$ 0.23 & 41.46 $\pm$ 0.15 \\
 & Base (w map) & 35.67 $\pm$ 0.76 & 45.28 $\pm$ 0.42 & 36.93 $\pm$ 0.61 & 41.46 $\pm$ 0.20 \\
 & Spatial Harness (w map) & 79.50 $\pm$ 1.00 & 92.56 $\pm$ 0.25 & 68.00 $\pm$ 1.06 & 84.22 $\pm$ 0.15 \\
\midrule
\multirow{3}{*}{Qwen3.5-27B} & Base (w/o map) & 77.17 $\pm$ 2.08 & 41.72 $\pm$ 0.75 & 59.47 $\pm$ 0.23 & 52.70 $\pm$ 0.81 \\
 & Base (w map) & 83.83 $\pm$ 1.61 & 60.94 $\pm$ 0.42 & 88.27 $\pm$ 0.23 & 71.81 $\pm$ 0.16 \\
 & Spatial Harness (w map) & 96.83 $\pm$ 0.29 & 98.06 $\pm$ 0.35 & 99.87 $\pm$ 0.23 & 98.25 $\pm$ 0.15 \\
\midrule
\multirow{3}{*}{Kimi K2.5} & Base (w/o map) & 69.67 $\pm$ 2.25 & 42.78 $\pm$ 1.06 & 54.40 $\pm$ 1.44 & 50.67 $\pm$ 1.06 \\
 & Base (w map) & 96.00 $\pm$ 0.87 & 75.83 $\pm$ 0.58 & 82.53 $\pm$ 0.61 & 81.27 $\pm$ 0.29 \\
 & Spatial Harness (w map) & 99.67 $\pm$ 0.29 & 95.33 $\pm$ 0.73 & 86.67 $\pm$ 1.40 & 94.10 $\pm$ 0.16 \\
\midrule
\multirow{3}{*}{GPT-5.4-mini} & Base (w/o map) & 28.83 $\pm$ 0.76 & 42.72 $\pm$ 0.59 & 60.27 $\pm$ 2.95 & 44.25 $\pm$ 1.10 \\
 & Base (w map) & 27.83 $\pm$ 1.04 & 42.61 $\pm$ 0.63 & 63.60 $\pm$ 1.44 & 44.79 $\pm$ 0.38 \\
 & Spatial Harness (w map) & 71.83 $\pm$ 1.61 & 92.44 $\pm$ 0.84 & 68.67 $\pm$ 0.61 & 82.86 $\pm$ 0.19 \\
\midrule
\multirow{3}{*}{GLM-5V-Turbo} & Base (w/o map) & 74.67 $\pm$ 0.29 & 56.44 $\pm$ 0.25 & 73.87 $\pm$ 1.40 & 64.06 $\pm$ 0.29 \\
 & Base (w map) & 77.83 $\pm$ 1.76 & 56.83 $\pm$ 0.67 & 84.67 $\pm$ 0.46 & 67.46 $\pm$ 0.69 \\
 & Spatial Harness (w map) & 90.17 $\pm$ 0.76 & 97.22 $\pm$ 0.10 & 71.20 $\pm$ 0.80 & 89.68 $\pm$ 0.27 \\
\bottomrule
\end{tabular*}
\end{table}

\subsection{Run-to-Run Variation}
\label{app:annotated-state-results}

Table~\ref{tab:mainresults-mindcube-map-seed} reports mean and standard deviation of accuracy over three runs in the GT-map setting.


\section{Case Studies}
\label{app:experimental-cases}
\label{app:exp-cases}

The following MindCube cases, both in the GT-map setting, give complete Reasoner trajectories. Each listing contains the code of every tool call, one block per turn; the text below it describes the Reasoner's reasoning and the engine responses, quoting the trajectory. Coordinates and directions follow the grid-map API, in which a yaw of $0$, $90$, $180$, and $270$ degrees faces right, up, left, and down on the map. Each case is compared with \textit{Base}, which receives the same images and GT map but no engine.

\begin{figure*}[!htb]
\centering
\begin{tcolorbox}[colback=white, colframe=black!70, boxrule=0.5pt, arc=0pt, left=6pt, right=6pt, top=3pt, bottom=3pt]
\textbf{Question:} Based on these four images (image 1, 2, 3, and 4) showing the red bottle from different viewpoints (front, left, back, and right), with each camera aligned with room walls and partially capturing the surroundings: From the viewpoint presented in image 4, what is to the left of the red bottle?\\
A.~Curtain \quad B.~Gray-green tufted backrest \quad C.~TV and electric fan \quad D.~Wall and door
\end{tcolorbox}
\vspace{2pt}
\begin{tikzpicture}[badge/.style={fill=blue!55!cyan, text=white, font=\small\bfseries, minimum size=4.5mm, inner sep=1pt, anchor=south west}]
\node[inner sep=0] (i1) at (0,0) {\includegraphics[width=0.243\textwidth]{figures/cases/among_group603_front_268.jpg}};
\node[inner sep=0, right=1.5mm of i1] (i2) {\includegraphics[width=0.243\textwidth]{figures/cases/among_group603_left_172.jpg}};
\node[inner sep=0, right=1.5mm of i2] (i3) {\includegraphics[width=0.243\textwidth]{figures/cases/among_group603_back_112.jpg}};
\node[inner sep=0, right=1.5mm of i3] (i4) {\includegraphics[width=0.243\textwidth]{figures/cases/among_group603_right_068.jpg}};
\foreach \i in {1,2,3,4} {\node[badge] at (i\i.south west) {\i};}
\end{tikzpicture}
\caption{The success case (item \texttt{among\_group603\_q3\_5\_1}). The top shows the question, and the bottom presents the four images, numbered as in the question. The reference answer is B.}
\label{fig:case-success}
\end{figure*}

\paragraph{Success: grounding the viewpoint change in the engine.}
Figure~\ref{fig:case-success} shows the question and the four images. The question asks: ``From the viewpoint presented in image 4, what is to the left of the red bottle?'' The options are A.~Curtain; B.~Gray-green tufted backrest; C.~TV and electric fan; D.~Wall and door. In the GT map, the red bottle is at $[5,5]$, the backrest at $[5,8]$, the curtain at $[2,5]$, the TV and fan at $[5,2]$, the wall and door at $[8,5]$, and the camera of image 4 at $[6,5]$ facing left. The trajectory of Qwen3.5-27B (item \texttt{among\_group603\_q3\_5\_1}) has two turns and one tool call:

\begin{lstlisting}[language=Python,basicstyle=\ttfamily\footnotesize,frame=single,breaklines=true,showstringspaces=false,columns=fullflexible]
# Turn 1
# Set camera to Image 4's position and facing
mujoco_sim.set_camera([6, 5], "left")
# Turn left to see what is to the left of the bottle from this viewpoint
mujoco_sim.turn_left()
# Capture the image to identify the object
image_path = mujoco_sim.capture_image()
\end{lstlisting}

\begin{figure}[!htb]
\centering
\includegraphics[width=0.6\textwidth]{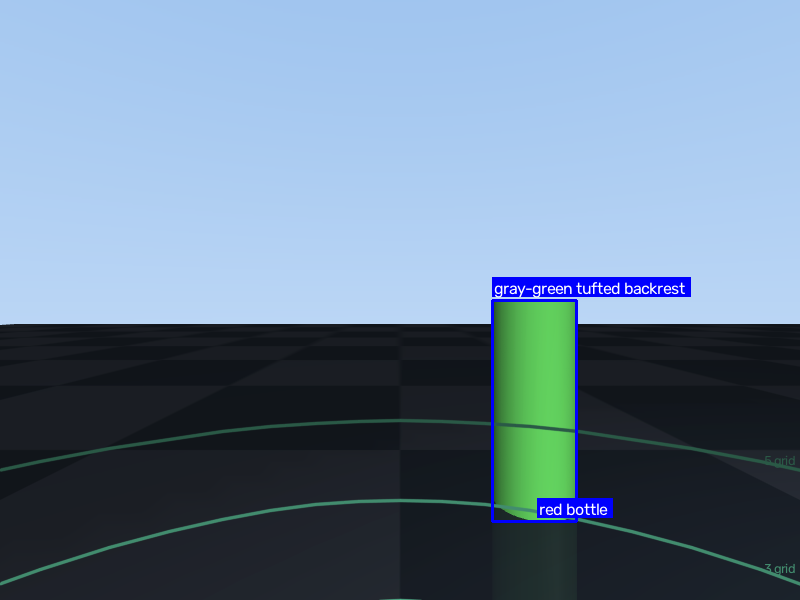}
\caption{View rendered by \texttt{capture\_image()} in the first turn of the success case.}
\label{fig:case-success-render}
\end{figure}

In the first turn, the Reasoner plans to ``set the camera to the position and facing of Image 4,'' ``turn the camera to the left to look at the area to the left of the bottle,'' and ``capture an image to identify the object in that direction.'' The engine renders the view facing down on the map (yaw $270$; Figure~\ref{fig:case-success-render}) and returns the gray-green tufted backrest at distance $3.2$ as the only entry of \texttt{objects\_in\_view}. In the second turn, the Reasoner calls no tool and answers B, stating that ``the tool captured the view after turning left and identified the `gray-green tufted backrest'.''

With the same images and GT map, \textit{Base}, which reasons directly over the same map without the engine, answers A in all three runs. In the first run, it reasons that ``from the camera's position at [6, 5] looking left towards the bottle at [5, 5], the direction `to the left' of the bottle corresponds to the direction the camera is facing (further decreasing x-values),'' and selects the curtain, the object straight ahead. The Reasoner answers B in all three runs. The viewpoint change is executed, not imagined: the engine renders the view after the turn, and the answer comes from this grounded feedback rather than from internal imagination.

\begin{figure*}[!htb]
\centering
\begin{tcolorbox}[colback=white, colframe=black!70, boxrule=0.5pt, arc=0pt, left=6pt, right=6pt, top=3pt, bottom=3pt]
\textbf{Question:} Based on these four images (image 1, 2, 3, and 4) showing the toy train from different viewpoints (front, left, back, and right), with each camera aligned with room walls and partially capturing the surroundings: From the viewpoint presented in image 3, what is to the left of the toy train?\\
A.~Window \quad B.~Printed glass door \quad C.~Black table \quad D.~Wall
\end{tcolorbox}
\vspace{2pt}
\begin{tikzpicture}[badge/.style={fill=blue!55!cyan, text=white, font=\small\bfseries, minimum size=4.5mm, inner sep=1pt, anchor=south west}]
\node[inner sep=0] (i1) at (0,0) {\includegraphics[width=0.243\textwidth]{figures/cases/among_group738_front_020.jpg}};
\node[inner sep=0, right=1.5mm of i1] (i2) {\includegraphics[width=0.243\textwidth]{figures/cases/among_group738_left_087.jpg}};
\node[inner sep=0, right=1.5mm of i2] (i3) {\includegraphics[width=0.243\textwidth]{figures/cases/among_group738_back_147.jpg}};
\node[inner sep=0, right=1.5mm of i3] (i4) {\includegraphics[width=0.243\textwidth]{figures/cases/among_group738_right_202.jpg}};
\foreach \i in {1,2,3,4} {\node[badge] at (i\i.south west) {\i};}
\end{tikzpicture}
\caption{The failure case (item \texttt{among\_group738\_q2\_5\_1}). The top shows the question, and the bottom presents the four images, numbered as in the question. The reference answer is A.}
\label{fig:case-failure}
\end{figure*}

\paragraph{Failure: overriding the engine's feedback.}
Figure~\ref{fig:case-failure} shows the question and the four images. The question asks: ``From the viewpoint presented in image 3, what is to the left of the toy train?'' The options are A.~Window; B.~Printed glass door; C.~Black table; D.~Wall. In the GT map, the toy train is at $[5,5]$, the black table at $[5,8]$, the wall at $[2,5]$, the printed glass door at $[5,2]$, the window at $[8,5]$, and the camera of image 3 at $[5,4]$ facing down. The trajectory of Kimi K2.5 (item \texttt{among\_group738\_q2\_5\_1}) has four turns and three tool calls:

\begin{lstlisting}[language=Python,basicstyle=\ttfamily\footnotesize,frame=single,breaklines=true,showstringspaces=false,columns=fullflexible]
# Turn 1
# Set camera to Image 3's position and facing
mujoco_sim.set_camera([5, 4], "down")
# From this viewpoint facing down, I need to check what's to the left of the toy train
# First, let me capture the current view to see the toy train
image_path = mujoco_sim.capture_image()
print("Objects in view from Image 3:", image_path)

# Turn 2
# From Image 3's position facing down, turn right to see what's to the left of the toy train
mujoco_sim.set_camera([5, 4], "down")
mujoco_sim.turn_right()  # This turns towards +x direction (window side)
image_path = mujoco_sim.capture_image()
print("Objects to the left of toy train from Image 3:", image_path)

# Turn 3
# Let me verify by checking both sides from Image 3's position
# First, reset to Image 3's position facing down
mujoco_sim.set_camera([5, 4], "down")

# Turn left to see what's on the other side of the toy train
mujoco_sim.turn_left()  # This turns towards +x direction (window side)
image_path_left = mujoco_sim.capture_image()
print("Objects to the right of toy train from Image 3:", image_path_left)

# Get distances to confirm
distances = mujoco_sim.get_distances()
print("Distances after turning left:", distances)
\end{lstlisting}

\begin{figure*}[!htb]
\centering
\begin{minipage}[t]{0.32\textwidth}
\centering
\includegraphics[width=\linewidth]{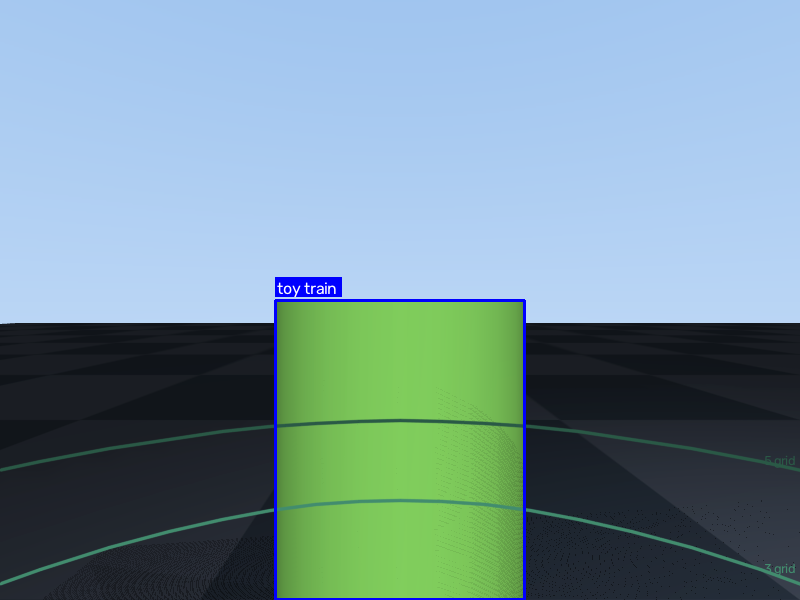}\\[1pt]
{\footnotesize Turn 1}
\end{minipage}
\hfill
\begin{minipage}[t]{0.32\textwidth}
\centering
\includegraphics[width=\linewidth]{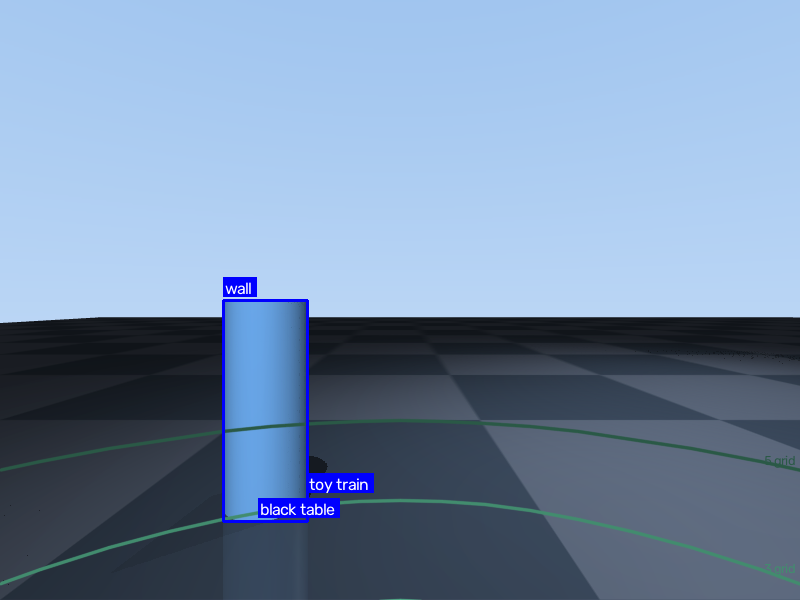}\\[1pt]
{\footnotesize Turn 2}
\end{minipage}
\hfill
\begin{minipage}[t]{0.32\textwidth}
\centering
\includegraphics[width=\linewidth]{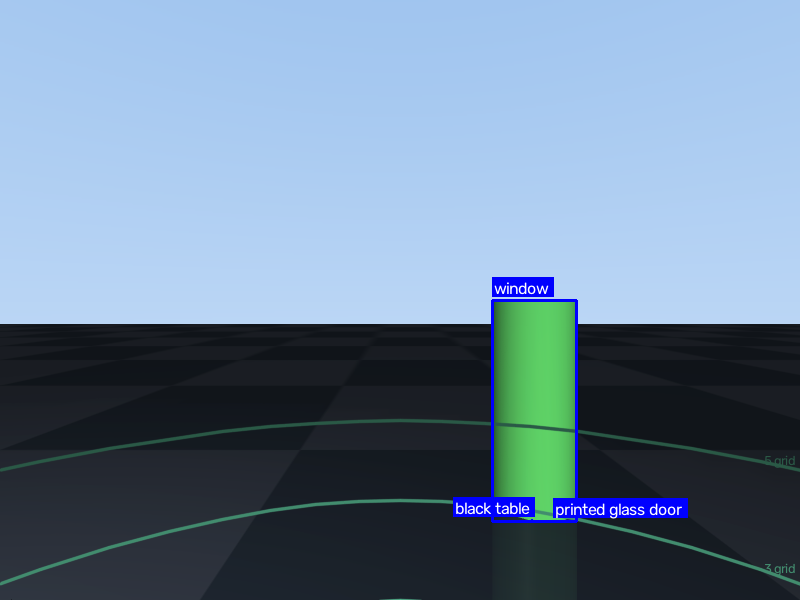}\\[1pt]
{\footnotesize Turn 3}
\end{minipage}
\caption{Views rendered by \texttt{capture\_image()} in the three tool-calling turns of the failure case, in order.}
\label{fig:case-failure-render}
\end{figure*}

In the first turn, the Reasoner reasons that when facing down, ``left = towards +x (right on grid),'' so ``the toy train's left would be towards [8, 5] where the window is,'' and captures the view of image 3 to verify; the engine returns the toy train at distance $1.0$ and the black table at $4.0$ (Figure~\ref{fig:case-failure-render}, left). In the second turn, it assumes that ``when facing down, right turn points towards +x'' and turns right; the engine renders the view facing left on the map (yaw $180$; Figure~\ref{fig:case-failure-render}, middle) and returns the wall at $3.2$. In the third turn, the Reasoner corrects the turn directions, ``turning right points towards -x'' and ``turning left points towards +x,'' but reinterprets the question: ``the question asks about the toy train's left, not the camera's left,'' and ``the toy train's left would be towards the camera's right (towards -x, which is the wall at [2, 5]),'' so ``the answer should be D. Wall.'' To double-check, it turns left and requests the distances; the engine renders the view facing right on the map (yaw $0$; Figure~\ref{fig:case-failure-render}, right), returns the window at $3.2$, and reports distances of $1.0$ to the toy train, $4.0$ to the black table, $3.162$ to the wall and the window, and $2.0$ to the printed glass door. In the fourth turn, the Reasoner calls no tool. It notes that ``when I turned left (towards +x direction), I saw the window at distance 3.2,'' but keeps that ``the toy train's left side is towards the -x direction (wall side)'' and answers D.

The reference answer is A. The engine executes every call correctly, and the capture after \texttt{turn\_left()} in the third turn returns the window. The Reasoner's analysis in the first turn also points to the window. In the third turn, however, the Reasoner reinterprets the question in the toy train's own frame, as if the train faced the camera, and keeps this reading in its final turn although it contradicts both its first analysis and the turn-left capture. The final answer coincides with that of \textit{Base} with the same map, which answers D in all three runs, whereas the Kimi K2.5 Reasoner answers A in its other two runs and the Qwen3.5-27B Reasoner answers A in all three. Grounded feedback shows the correct answer but does not force the model to use it: this failure occurs when the Reasoner sets aside the engine's observation in favor of its internal interpretation of the question.

\clearpage
\newpage
\clearpage
\section*{NeurIPS Paper Checklist}

\begin{enumerate}

\item {\bf Claims}
    \item[] Question: Do the main claims made in the abstract and introduction accurately reflect the paper's contributions and scope?
    \item[] Answer: \answerYes{} 
    \item[] Justification: The abstract and Section~\ref{sec:introduction} state the claims: perception is made explicit as a world state, and reasoning is grounded in feedback computed by a physics engine. Section~\ref{sec:experiments} supports them on four benchmarks with open- and closed-weight VLMs.

\item {\bf Limitations}
    \item[] Question: Does the paper discuss the limitations of the work performed by the authors?
    \item[] Answer: \answerYes{} 
    \item[] Justification: Please see Section~\ref{sec:discuss}.

\item {\bf Theory assumptions and proofs}
    \item[] Question: For each theoretical result, does the paper provide the full set of assumptions and a complete (and correct) proof?
    \item[] Answer: \answerNA{} 
    \item[] Justification: The paper does not include theoretical results.

    \item {\bf Experimental result reproducibility}
    \item[] Question: Does the paper fully disclose all the information needed to reproduce the main experimental results of the paper to the extent that it affects the main claims and/or conclusions of the paper (regardless of whether the code and data are provided or not)?
    \item[] Answer: \answerYes{} 
    \item[] Justification: Section~\ref{subsec:exp-setting} and Appendix~\ref{app:implementation-details} give the benchmarks, models, prompts, and evaluation protocol, and the code is released at \url{https://github.com/tmlr-group/Spatial-Harness}.

\item {\bf Open access to data and code}
    \item[] Question: Does the paper provide open access to the data and code, with sufficient instructions to faithfully reproduce the main experimental results, as described in supplemental material?
    \item[] Answer: \answerYes{} 
    \item[] Justification: The code is publicly available at \url{https://github.com/tmlr-group/Spatial-Harness}, and Appendix~\ref{app:implementation-details} describes the implementation and evaluation setup.

\item {\bf Experimental setting/details}
    \item[] Question: Does the paper specify all the training and test details (e.g., data splits, hyperparameters, how they were chosen, type of optimizer) necessary to understand the results?
    \item[] Answer: \answerYes{} 
    \item[] Justification: Section~\ref{subsec:exp-setting} gives the experimental settings, and Appendix~\ref{app:implementation-details} gives the implementation, decoding, and scoring details.

\item {\bf Experiment statistical significance}
    \item[] Question: Does the paper report error bars suitably and correctly defined or other appropriate information about the statistical significance of the experiments?
    \item[] Answer: \answerYes{} 
    \item[] Justification: Please see Table~\ref{tab:mainresults-mindcube-map-seed}.

\item {\bf Experiments compute resources}
    \item[] Question: For each experiment, does the paper provide sufficient information on the computer resources (type of compute workers, memory, time of execution) needed to reproduce the experiments?
    \item[] Answer: \answerYes{} 
    \item[] Justification: Appendix~\ref{app:compute} gives the GPUs used for the locally served models; the closed-weight models are accessed through APIs.

\item {\bf Code of ethics}
    \item[] Question: Does the research conducted in the paper conform, in every respect, with the NeurIPS Code of Ethics \url{https://neurips.cc/public/EthicsGuidelines}?
    \item[] Answer: \answerYes{} 
    \item[] Justification: Our paper conforms to the NeurIPS Code.

\item {\bf Broader impacts}
    \item[] Question: Does the paper discuss both potential positive societal impacts and negative societal impacts of the work performed?
    \item[] Answer: \answerYes{} 
    \item[] Justification: Please see Section~\ref{app:impact}.

\item {\bf Safeguards}
    \item[] Question: Does the paper describe safeguards that have been put in place for responsible release of data or models that have a high risk for misuse (e.g., pre-trained language models, image generators, or scraped datasets)?
    \item[] Answer: \answerNA{} 
    \item[] Justification: The paper does not have such risks.

\item {\bf Licenses for existing assets}
    \item[] Question: Are the creators or original owners of assets (e.g., code, data, models), used in the paper, properly credited and are the license and terms of use explicitly mentioned and properly respected?
    \item[] Answer: \answerYes{} 
    \item[] Justification: For the datasets and open-sourced models that we use in our experiments, we cited the papers that produced them.

\item {\bf New assets}
    \item[] Question: Are new assets introduced in the paper well documented and is the documentation provided alongside the assets?
    \item[] Answer: \answerYes{} 
    \item[] Justification: The code is released at \url{https://github.com/tmlr-group/Spatial-Harness} with a README that documents installation, configuration, and evaluation, and Appendix~\ref{app:implementation-details} gives the implementation details.

\item {\bf Crowdsourcing and research with human subjects}
    \item[] Question: For crowdsourcing experiments and research with human subjects, does the paper include the full text of instructions given to participants and screenshots, if applicable, as well as details about compensation (if any)?
    \item[] Answer: \answerNA{} 
    \item[] Justification: This paper does not include crowdsourcing.

\item {\bf Institutional review board (IRB) approvals or equivalent for research with human subjects}
    \item[] Question: Does the paper describe potential risks incurred by study participants, whether such risks were disclosed to the subjects, and whether Institutional Review Board (IRB) approvals (or an equivalent approval/review based on the requirements of your country or institution) were obtained?
    \item[] Answer: \answerNA{} 
    \item[] Justification: The paper does not include crowdsourcing.

\item {\bf Declaration of LLM usage}
    \item[] Question: Does the paper describe the usage of LLMs if it is an important, original, or non-standard component of the core methods in this research? Note that if the LLM is used only for writing, editing, or formatting purposes and does \emph{not} impact the core methodology, scientific rigor, or originality of the research, declaration is not required.
    \item[] Answer: \answerYes{} 
    \item[] Justification: VLMs serve as the Perceiver and the Reasoner, and an LLM judge scores free-form answers. Section~\ref{sec:methodology}, Section~\ref{subsec:exp-setting}, and Appendix~\ref{app:llm-usage} describe the models and how they are used.

\end{enumerate}

\end{document}